\documentclass[letterpaper, 10pt, conference]{ieeeconf}  %

\IEEEoverridecommandlockouts                              %

\makeatletter
\let\NAT@parse\undefined
\makeatother

\usepackage[usenames,dvipsnames]{xcolor}
\usepackage{soul}

\usepackage{xspace}
\newcommand{\acronym}{Statler\xspace}
\newcommand{\vanillallm}{Code-as-Policies\xspace}

\usepackage[textwidth=1.5cm, disable]{todonotes}

\newcommand{\note}[4][]{{\todo[author=#2,color=#3,size=\scriptsize,fancyline,caption={},#1]{#4}}}

\newcommand{\Ben}[2][]{{\note[#1]{Ben}{brown!20}{#2}}}

\newcommand{\response}[1]{\vspace{3pt}\hrule\vspace{3pt}\textbf{#1:}}

\newcommand{\cutforspace}[1]{}

\usepackage[utf8]{inputenc} %
\usepackage[T1]{fontenc}    %
\usepackage[square, numbers, sort&compress]{natbib}

\usepackage{caption}
\usepackage[bookmarks=true,colorlinks=true,citecolor=blue,linkcolor=blue]{hyperref}
\usepackage{url}            %
\usepackage{booktabs}       %
\usepackage{amsfonts}       %
\usepackage{nicefrac}       %
\usepackage{microtype}      %
\usepackage{subcaption}
\usepackage{bm}
\usepackage{wrapfig}

\usepackage{tabularx}
\usepackage{multicol}

\newcolumntype{Y}{>{\centering\arraybackslash}X}

\usepackage{authblk}
\usepackage{microtype}
\usepackage{setspace}
\usepackage{flushend}

\usepackage{float}
\newfloat{fixed}{H}{lop}
\floatname{fixed}{Fixed Environment}

\usepackage{newfloat}
\DeclareFloatingEnvironment{codefloat}

\newcommand{\codefont}{\fontfamily{lmtt}\selectfont}
\usepackage{listings}

\usepackage{parcolumns}
\lstdefinestyle{datalogstyle}{
	basicstyle={\codefont\scriptsize},  %
	xleftmargin={6pt},
    xrightmargin={6pt},
	numbers=left,
    columns=flexible,
    breakindent=0pt,
    breaklines=true, 
	frame=tb,
	stepnumber=1,
	firstnumber=1,
	numberfirstline=true,
	tabsize=2,
	extendedchars=true,
	breaklines=true,
	columns=fullflexible,
	keepspaces=true,
	escapeinside={@}{@},
	firstnumber=last,
	captionpos=b, 
	commentstyle=\color{black!65},
	numberstyle=\tiny\color{black!65},
	stringstyle=\color{codepurple},
	breakatwhitespace=false, 
	keepspaces=true,              
    mathescape=true, 
	numbersep=5pt,                  
	showspaces=false,                
	showstringspaces=false,
	showtabs=false,
	aboveskip={0.8\baselineskip},
	belowskip={0.2\baselineskip},
}
\definecolor{aigold}{RGB}{244,210, 1} 
\definecolor{aigreen}{RGB}{213, 245, 227}

\definecolor{humanpurple}{RGB}{235, 222, 240}

\definecolor{commentgray}{RGB}{86, 101, 115}

\definecolor{aired}{RGB}{255,180,181}

\usepackage[precision=2, unit=mm]{lengthconvert}
\usetikzlibrary{patterns,calc,fit,math}
\tikzset{
    response/.style 2 args = {
        draw, line width=1pt, inner sep=0, outer sep=0,
        fit=(#1) (#2)}
}

\newcommand{\greencheck}{}%
\DeclareRobustCommand{\greencheck}{%
  \tikz\fill[scale=0.3, color=green!50!black]
  (0,.35) -- (.25,0) -- (1,.7) -- (.25,.15) -- cycle;%
}

\newcommand{\redx}{}%
\DeclareRobustCommand{\redx}{%
    \tikz[scale=0.23, color=red!50!black] {
    \draw[line width=0.7,line cap=round] (0,0) to [bend left=6] (1,1);
    \draw[line width=0.7,line cap=round] (0.2,0.95) to [bend right=3] (0.8,0.05);}
}

\newcommand{\nfrac}[2]{$#1$~\raisebox{1.2pt}{\scalebox{0.75}{$#2$}}}
\newcommand{\nfracb}[2]{$\bm{#1}$~\raisebox{1.2pt}{\scalebox{0.75}{$\bm{#2}$}}}

\usepackage[noabbrev,capitalize]{cleveref} %
\crefname{equation}{Equation}{Equations}   %
\crefname{section}{Section}{Sections}      %
\crefname{footnote}{Footnote}{Footnotes}   
\crefname{listing}{Prompt}{Prompts}
\crefname{assumption}{Assumption}{Assumptions}
\crefname{line}{Line}{Lines}   %

\usepackage{xpatch}
\usepackage[compact]{titlesec}
\usepackage{setspace}
\AtBeginDocument{%
  \addtolength\abovedisplayskip{-0.25\baselineskip}%
  \addtolength\belowdisplayskip{-0.25\baselineskip}%
  \addtolength\abovedisplayshortskip{-0.25\baselineskip}%
  \addtolength\belowdisplayshortskip{-0.25\baselineskip}%
}

\title{\LARGE \bf
Statler: \underline{Stat}e-Maintaining \underline{L}anguage Models for \underline{E}mbodied \underline{R}easoning
}

\author[1]{Takuma Yoneda$^*$\thanks{TTIC}}%
\author[1]{Jiading Fang$^*$}%
\author[2]{Peng Li$^*$}%
\author[3]{Huanyu Zhang$^*$}%
\author[3]{Tianchong Jiang}%
\author[1]{Shengjie Lin}%
\author[3]{\\Ben Picker}%
\author[1]{David Yunis}%
\author[1]{Hongyuan Mei}%
\author[1]{Matthew R. Walter}%
\affil[1]{Toyota Technological Institute at Chicago%
\protect\\[-3pt]%
\texttt{\{takuma,fjd,slin,dyunis,hongyuan,mwalter\}@ttic.edu}}
\affil[2]{Fudan University%
\protect\\[-3pt]%
\texttt{lip21@m.fudan.edu.cn}}
\affil[3]{University of Chicago%
\protect\\[-3pt]%
\texttt{\{huanyu,tianchongj,bpicker\}@uchicago.edu}}%

\begin{document}

\twocolumn[{%
    \begin{@twocolumnfalse}
        \maketitle
        \begin{center}
            \vspace{-15pt}
            \centering
            \captionsetup{type=figure}
            \includegraphics[width=0.85\textwidth]{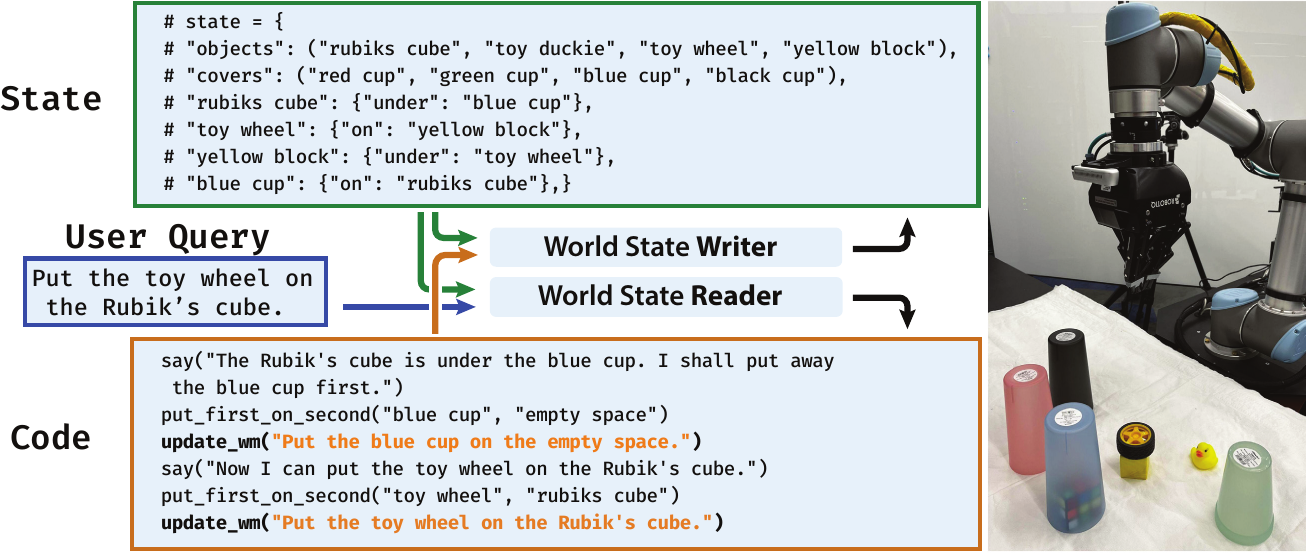}
            \captionof{figure}{Our \acronym framework enables robots to carry out complex tasks specified in natural language that require reasoning over long time horizons. Integral to our model are its world-state writer and world-state reader, two instances of general LLMs responsible for maintaining the explicit world state and generating code that enables the robot to carry out the task.} \label{fig:diagram}
        \end{center}
    \end{@twocolumnfalse}
}]

\begin{abstract}
    There has been a significant research interest in employing large language models to empower intelligent robots with complex reasoning. 
    Existing work focuses on harnessing their abilities to reason about the histories of their actions and observations. 
    In this paper, we explore a new dimension in which large language models may benefit robotics planning. 
    In particular, we propose \acronym, a framework in which large language models are prompted to maintain an estimate of the world state, which are often {unobservable}, and track its transition as new actions are taken. %
    Our framework then conditions each action on the estimate of the current world state. %
    Despite being conceptually simple, our \acronym framework significantly outperforms strong competing methods (e.g., Code-as-Policies) on several robot planning tasks. 
    Additionally, it has the potential advantage of scaling up to more challenging long-horizon planning tasks. We release our code \href{https://github.com/ripl/statler}{here}.
\end{abstract}
\def\thefootnote{*}\footnotetext{Denotes equal contribution; Contribution by each author can be found in the appendix.}\def\thefootnote{\arabic{footnote}}

\begin{figure*}[t]
\begin{minipage}[t]{\linewidth}
\begin{minipage}[t]{0.32\linewidth}
\begin{lstlisting}[caption={The prompt and \sethlcolor{aigreen}\hl{desired output} of a vanilla LLM.},label={lst:cup-ball-vanilla-prompt},firstnumber=auto]
@
\textcolor{commentgray}{\# Initial state}
@
cups = [False, True, False] @\label{line:init-cup-state}@
Swapping cup 1 with cup 2
Swapping cup 0 with cup 2
Swapping cup 1 with cup 2
cups = @\sethlcolor{aigreen}\hl{[True, False, False]}@ @\label{line:gptout}@  
@ @
@ @
\end{lstlisting}
\end{minipage}
\hfill
\begin{minipage}[t]{0.32\linewidth}
\begin{lstlisting}[caption={The prompt and \sethlcolor{aigreen}\hl{desired output} of LLM+CoT.},label={lst:cot-cup-ball-prompt},firstnumber=auto]
@
\textcolor{commentgray}{\# Initial state}
@
cups = [False, True, False] @\label{line:init-cup-state}@
Swapping cup 1 with cup 2
Swapping cup 0 with cup 2
Swapping cup 1 with cup 2
cups = @\sethlcolor{aigreen}\hl{[False, False, True]}@
cups = @\sethlcolor{aigreen}\hl{[True, False, False]}@
cups = @\sethlcolor{aigreen}\hl{[True, False, False]}@ @\label{line:gptout}@ 
\end{lstlisting}
\end{minipage}
\hfill
\begin{minipage}[t]{0.32\linewidth}
\begin{lstlisting}[caption={The prompt and \sethlcolor{aigreen}\hl{desired output} of LLM+State.},label={lst:state-maintaining-cup-ball-prompt},firstnumber=auto]
@
\textcolor{commentgray}{\# Initial state}
@
cups = [False, True, False] @\label{line:init-cup-state}@
Swapping cup 1 with cup 2
cups = @\sethlcolor{aigreen}\hl{[False, False, True]}@
Swapping cup 0 with cup 2
cups = @\sethlcolor{aigreen}\hl{[True, False, False]}@
Swapping cup 1 with cup 2
cups = @\sethlcolor{aigreen}\hl{[True, False, False]}@  
\end{lstlisting}
\end{minipage}
\end{minipage}
\vspace{-16pt}
\end{figure*}
\section{Introduction}
Large language models (LLMs) exhibit strong reasoning capabilities that are harnessed to perform a wide range of downstream tasks such as dialogue and code generation~\cite{Kojima2022LargeLM,
DBLP:journals/corr/abs-2107-03374,DBLP:journals/corr/abs-2303-08774}. 
The robotics community has recently seen a significant interest in empowering robots with LLMs, enabling them to understand natural language commands and perform tasks that require sophisticated reasoning~\cite{Ahn2022DoAI,Liang2022CodeAP, Huang2022InnerME,yang2023foundation}. 
However, existing methods are model-free: they use LLMs as policy functions that generate future actions only conditioned on previous actions and observations. %

In this paper, we propose a simple yet effective model-based approach. 
Our framework---named \acronym---maintains a running estimate of the world \underline{stat}e by prompting large \underline{l}anguage models and performs multistep \underline{e}mbodied \underline{r}easoning conditioned on the estimated state. 
\cref{fig:diagram} illustrates this framework. 
In particular, \acronym utilizes a pair of prompted LLMs: instructed by a few demonstrations, the \textbf{world-state reader} takes as input the user query, reads the estimated world state, and generates an executable action (e.g, a code snippet); instructed by another set of demonstrations, the \textbf{world-state writer} updates the world state estimate based on the action. 
This mechanism resembles how a domain-specific formal language tracks a symbolic world state~\cite{Nordmann2014ASO}, but enjoys greater flexibility since pretrained large language models are known to be domain-agnostic. As we will see soon in \cref{sec:experiments}, the prompts in our experiments are generic and users of our framework will have minimal design workload. 

Our \acronym framework is primarily inspired by classical model-based reinforcement learning. 
In a model-based approach, an environment (or world) model learns to capture the dynamics of the environment (e.g., possible outcomes of an action) so that the policy conditioned on the model state will take more informed actions~\citep{sutton98}.
In our framework, the LLMs have acquired massive amounts of commonsense knowledge from pretraining, and they are elicited---by a few demonstrations---to behave as an environment model, estimating the world state and facilitating decision making. 

Another motivation of our design is to handle missing data. 
In robotics tasks, we often have to cope with latent world dynamics that are not directly observable.
In such scenarios, explicitly maintaining an estimated world state improves decision making, although the estimates might be imperfect. 
This is analogous to graphical models with latent variables: spelling out latent variables and imputing their values is often helpful for reasoning about the target variables, although the imputation may not be perfect~\citep{koller2009probabilistic}. %

The final motivation of our state-maintaining design is its potential to scale up to long-horizon planning tasks. 
In multistep reasoning and planning, an LLM has to implicitly maintain the world state in its internal representation, which has been demonstrated to be difficult in previous work~\citep{Anthropic,nelson23,Sun2021DoLL,valmeekam2023planning,guan2023leveraging,liu2023llm+, silver2023generalized}. 
By explicitly maintaining an estimated world state, our framework makes it easier to track and consult the world state at any step of reasoning and planning, thus carrying a higher chance of success in long-horizon tasks. 

In the following sections, we will show that our framework performs as expected: 
in \cref{sec:motivation}, we demonstrate the concept with a pedagogical example; 
in \cref{sec:method}, we introduce the \acronym framework; in \cref{sec:experiments}, we present the experiments, in which our framework significantly outperforms strong competing methods such as Code-as-Policies~\cite{Liang2022CodeAP}.

\section{Motivating Example}\label{sec:motivation}
We use \textit{three-cups-and-a-ball}, a simple shell game, to demonstrate the effectiveness of our state-maintaining idea. 
In this game, a ball is covered under one of three identical cups and the initial position of the ball is known to the player. 
In each of $K$ rounds, we randomly swap two cups' positions. Finally, we ask the player to guess the position of the ball. 

We present three separate cases of using LLMs to play this game using GPT-3 (precisely, text-davinci-003). \cref{lst:cup-ball-vanilla-prompt} demonstrates the simplest case: We represent the state with Boolean variables with \texttt{True} indicating ``ball is here''. We feed the initial state and the $K$ rounds of swaps into GPT-3, instructing it to complete the final state.
\cref{lst:cot-cup-ball-prompt} is an improved way: after reading $K$ rounds of swaps, GPT-3 is asked to give all the intermediate states over the game. 
This version is inspired by Chain-of-Thought prompting~\citep{DBLP:conf/nips/Wei0SBIXCLZ22}, which improves the performance of an LLM by requesting it to spell out its intermediate reasoning steps. Finally, 
\cref{lst:state-maintaining-cup-ball-prompt} is a simple instantiation of our state-maintaining idea: we ask GPT-3 to return the current state immediately after reading each round of swaps, stimulating the model to track and update the state as the game progresses. 

\begin{figure}[!t]
    \centering
    \includegraphics[width=0.9\linewidth]{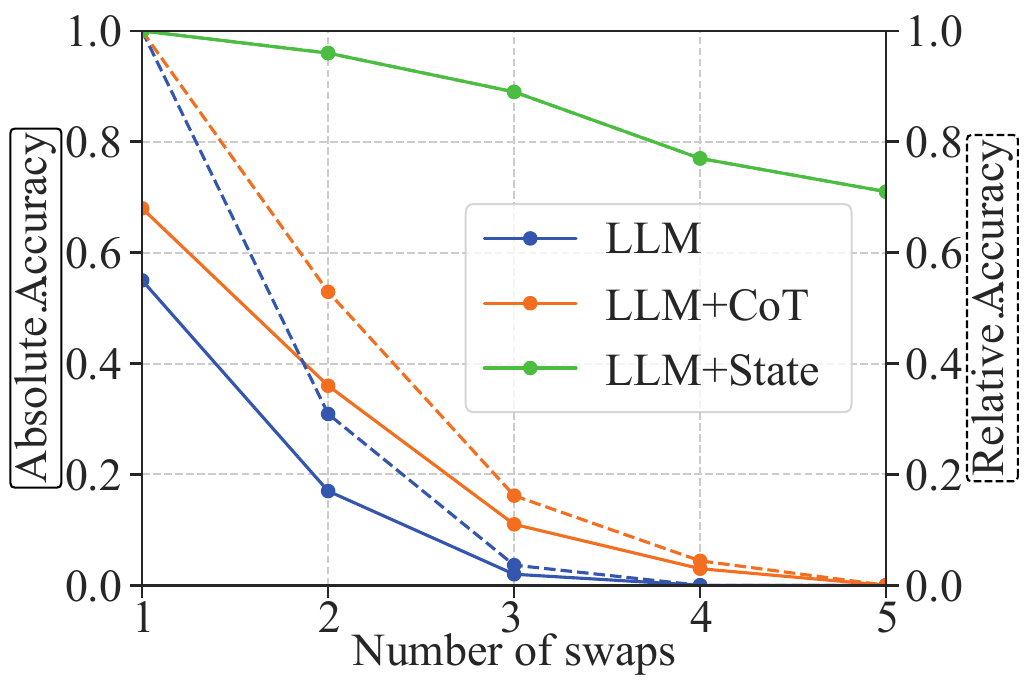}
    \caption{Model accuracies on the \emph{three-cups-and-a-ball} shell game. LLM+State is a simplified version of our proposed \acronym framework. 
    For each method, the solid line shows how its accuracy $a(n)$ changes with the number of swaps $n$. 
    The dashed line is the \emph{relative} accuracy: $r(n) = a(n)/a(1)$. Intuitively, it measures how fast the performance decreases from a \emph{hypothetically perfect} one-swap performance. Note that LLM+State indeed achieves $a(1)=100\%$
    } \label{fig:shell-game-plot}
\end{figure}

We evaluate these methods with a range of $K$; for each $K$ and each method, 
we feed $30$ demonstrations with various numbers of swaps to the model, and repeat the experiment $100$ times. 
\cref{fig:shell-game-plot} visualizes the average accuracies. The state-maintaining method significantly outperforms the other methods, with the performance gap increasing with $K$.

\section{Method}\label{sec:method}%
\begin{figure}[!t]
    \centering
    \begin{subfigure}{0.485\textwidth}
        \centering
        \begin{tikzpicture}[utterance/.style={rectangle, rounded corners, draw=black, font=\scriptsize, minimum width=5.75cm, minimum height=0.5cm, text width = 5.8cm, fill=black!10, outer sep=0pt},
            empty/.style={outer sep=1pt, inner sep=0pt}]
            \node [empty, label=below:{\footnotesize Initial State}] (before) at (0,0) {\includegraphics[width=0.32\linewidth]{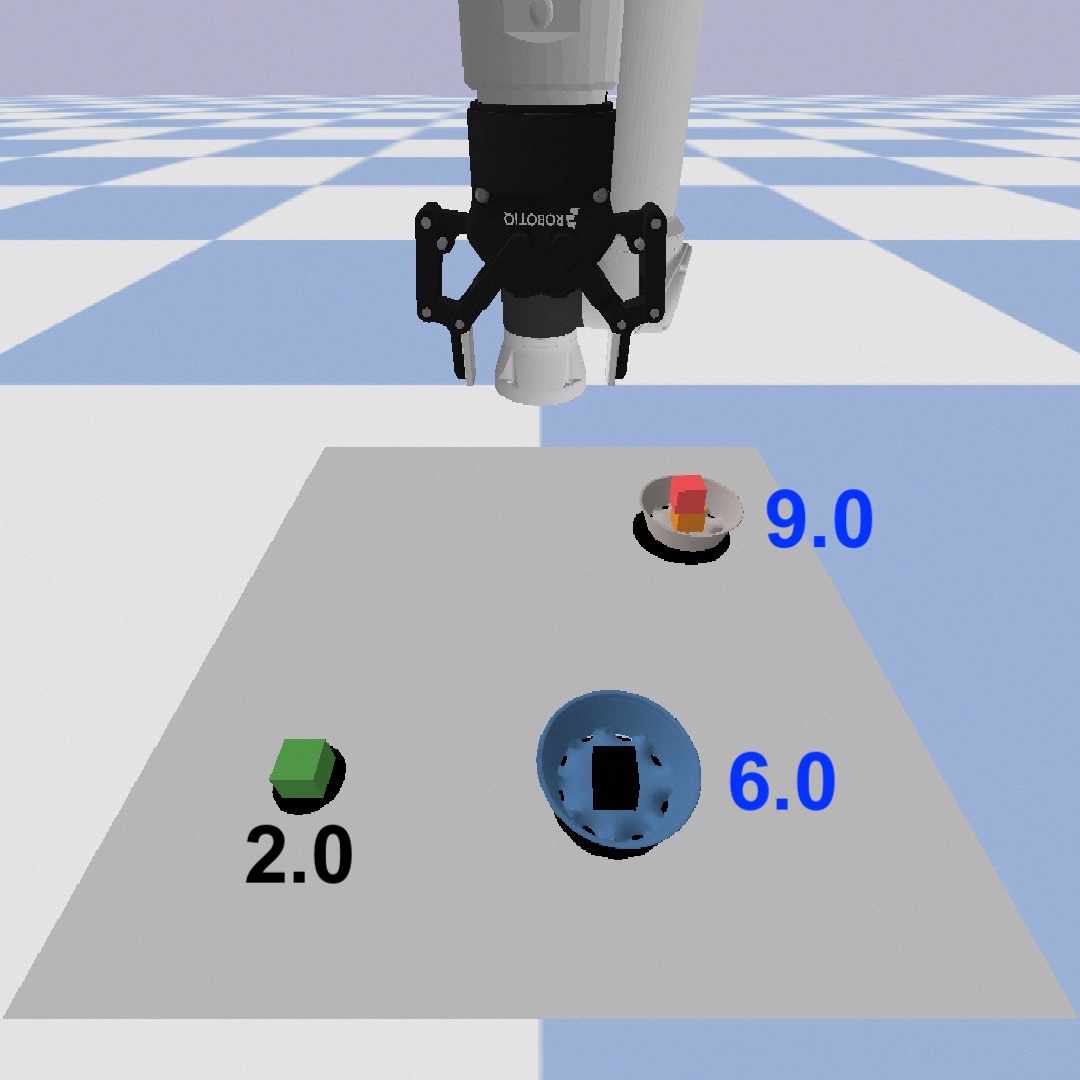}};
            \node [empty, right=0 of before, label=below:{\footnotesize \vanillallm}] (after-cap) {\includegraphics[width=0.32\linewidth]{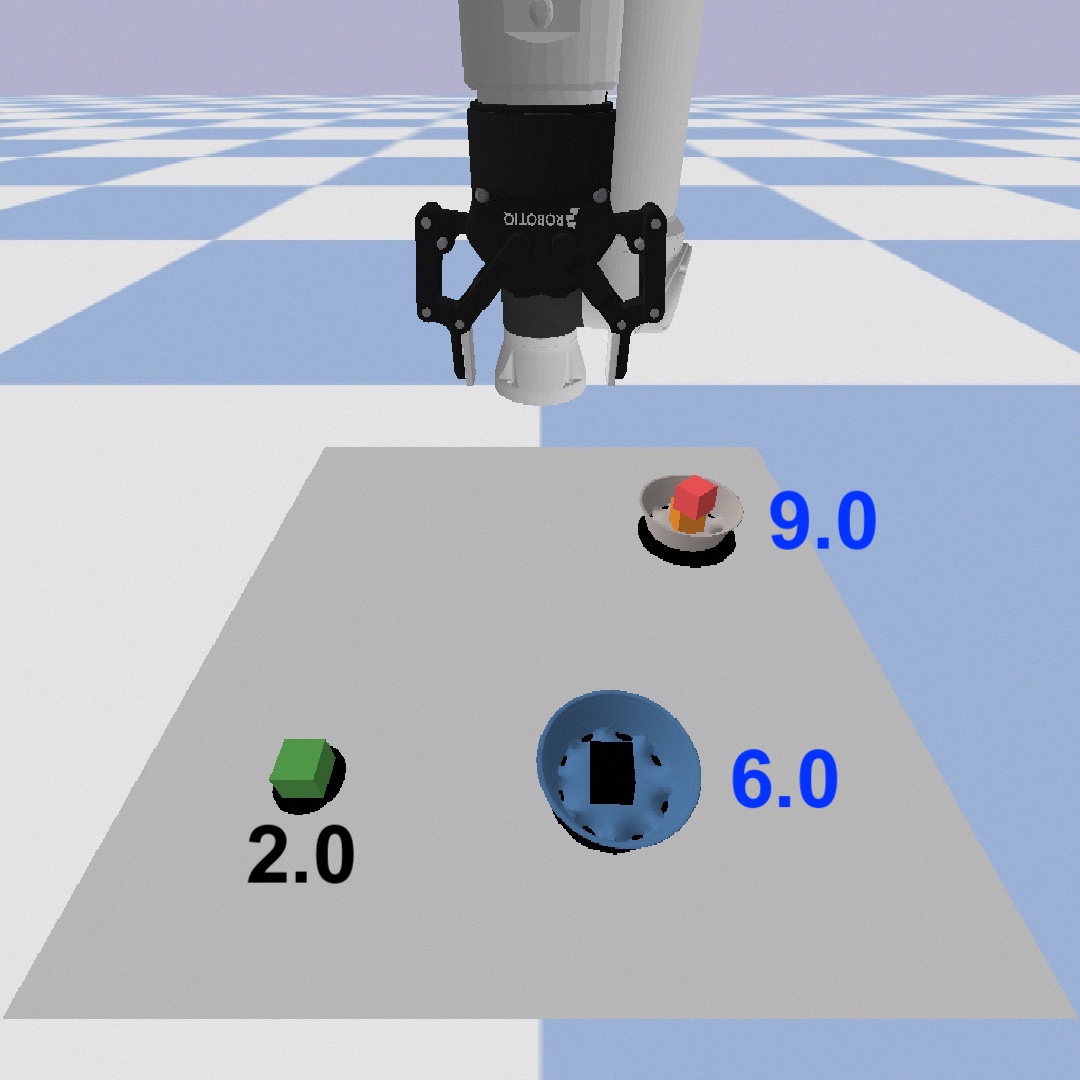}};
            \node [empty, right=0 of after-cap, label=below:{\footnotesize \acronym (ours)}] (after-ours) {\includegraphics[width=0.32\linewidth]{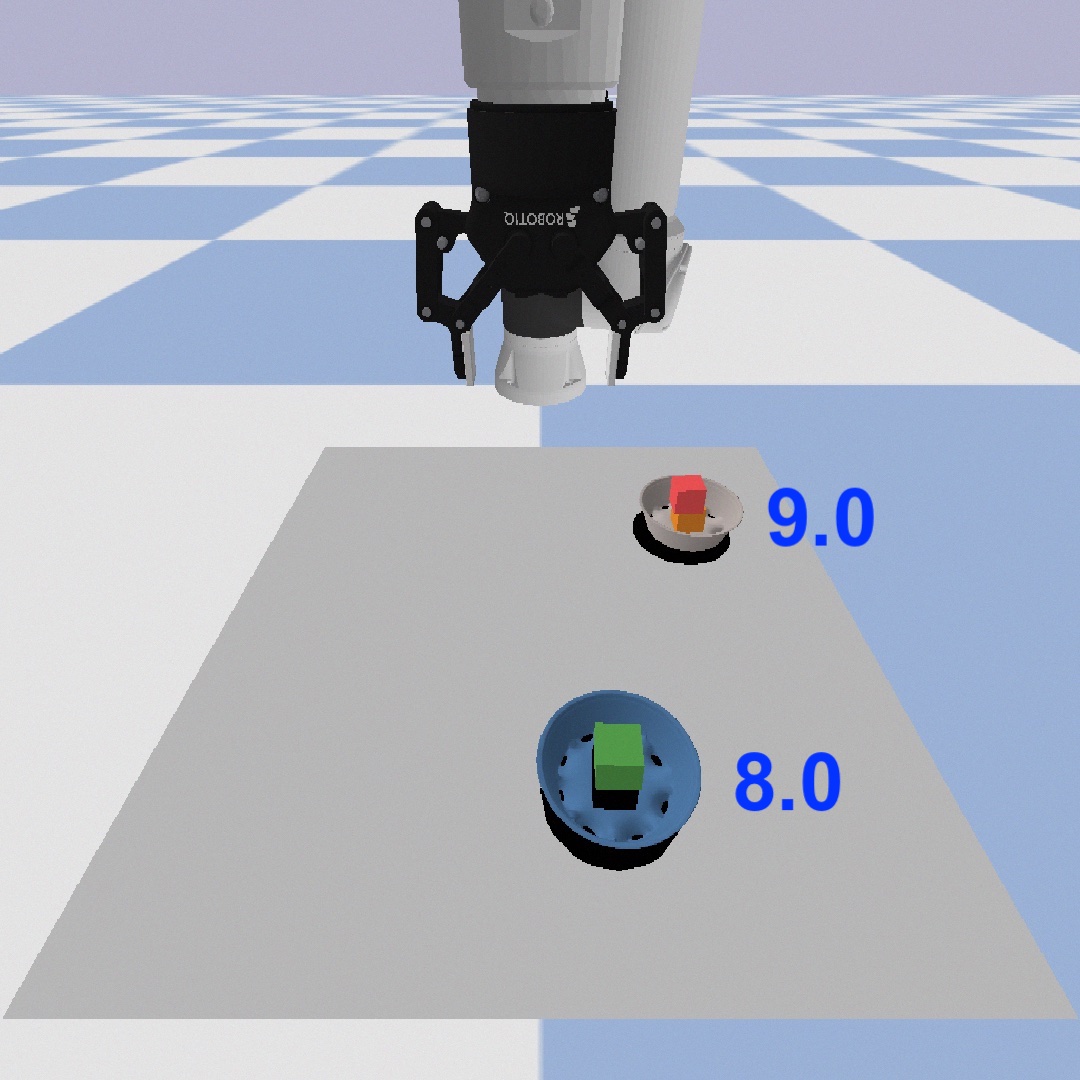}};
            \node[empty, above=0 and 0 of before.north west,anchor=south west] (speak) {\includegraphics[height=5mm]{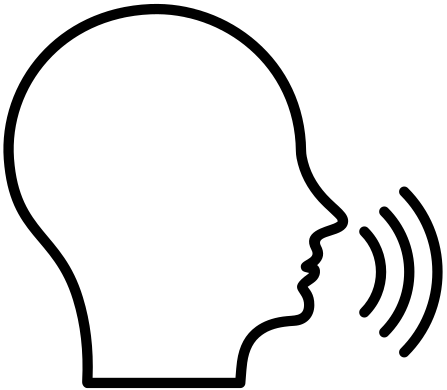}};
            \node[utterance, right= 1pt and 0 of speak.south east, anchor=south west, rounded corners] (utterance) {\textls[-28]{put a block in the blue bowl so that the total weight of blocks in the blue bowl is less than what is in the gray bowl}};
            \node [circle,draw=black, fill=white, inner sep=1pt, minimum size=5pt,  above right=5pt of after-ours.south west, anchor=south west] (check) {\greencheck};
            \node [circle,draw=black, fill=white, inner sep=1pt, minimum size=5pt, above right=5pt of after-cap.south west, anchor=south west] (check) {\redx};
        \end{tikzpicture}
    \end{subfigure}\\
    \begin{subfigure}{0.485\textwidth}
        \centering
        \begin{tikzpicture}[utterance/.style={rectangle, rounded corners, draw=black, font=\scriptsize, minimum width=5.75cm, minimum height=0.5cm, text width = 5.75cm, fill=black!10, outer sep=0pt},
            empty/.style={outer sep=1pt, inner sep=0pt}]
            \node [empty, label=below:{\footnotesize Initial State}] (before) at (0,0) {\includegraphics[width=0.32\linewidth]{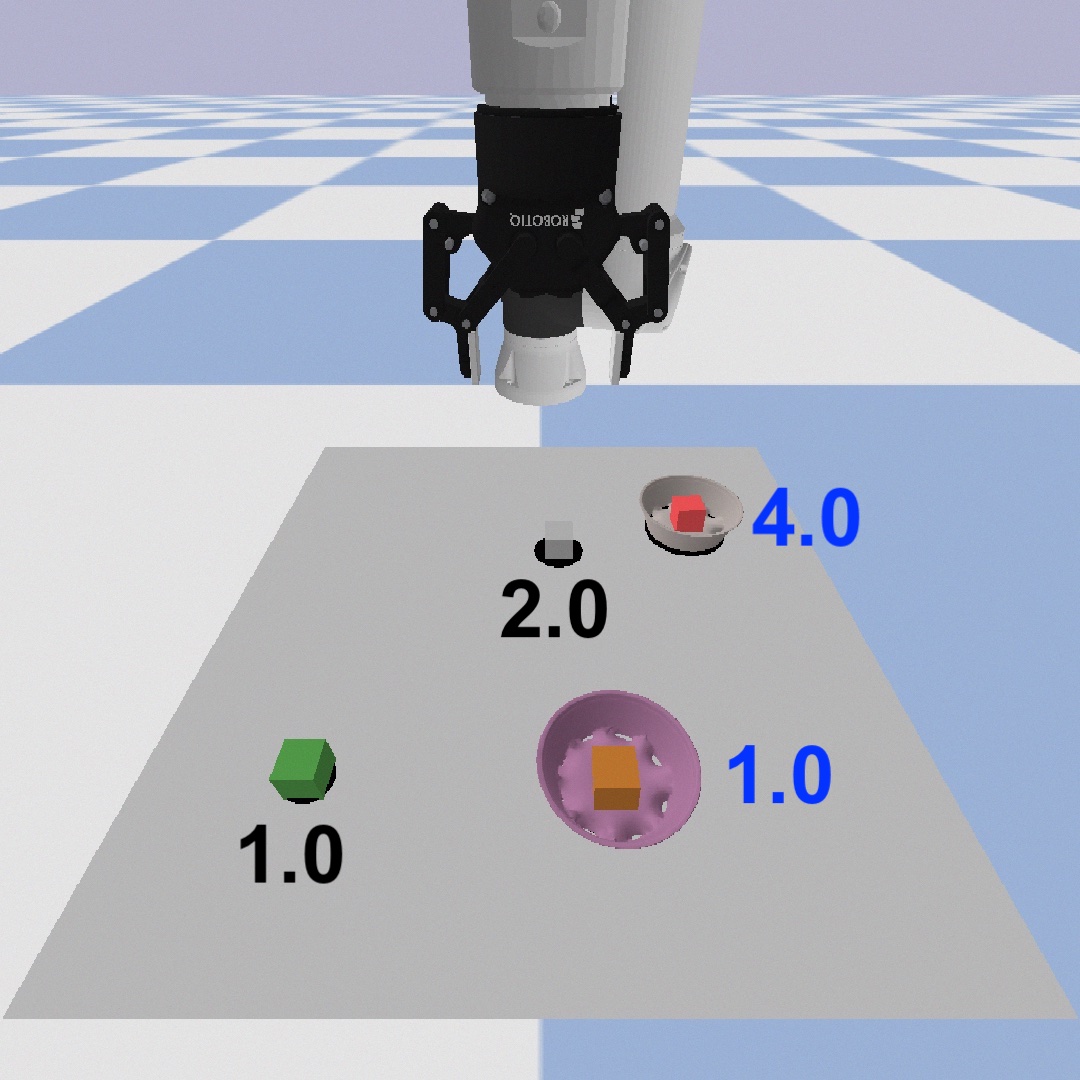}};
            \node [empty, right=0 of before, label=below:{\footnotesize \vanillallm}] (after-cap) {\includegraphics[width=0.32\linewidth]{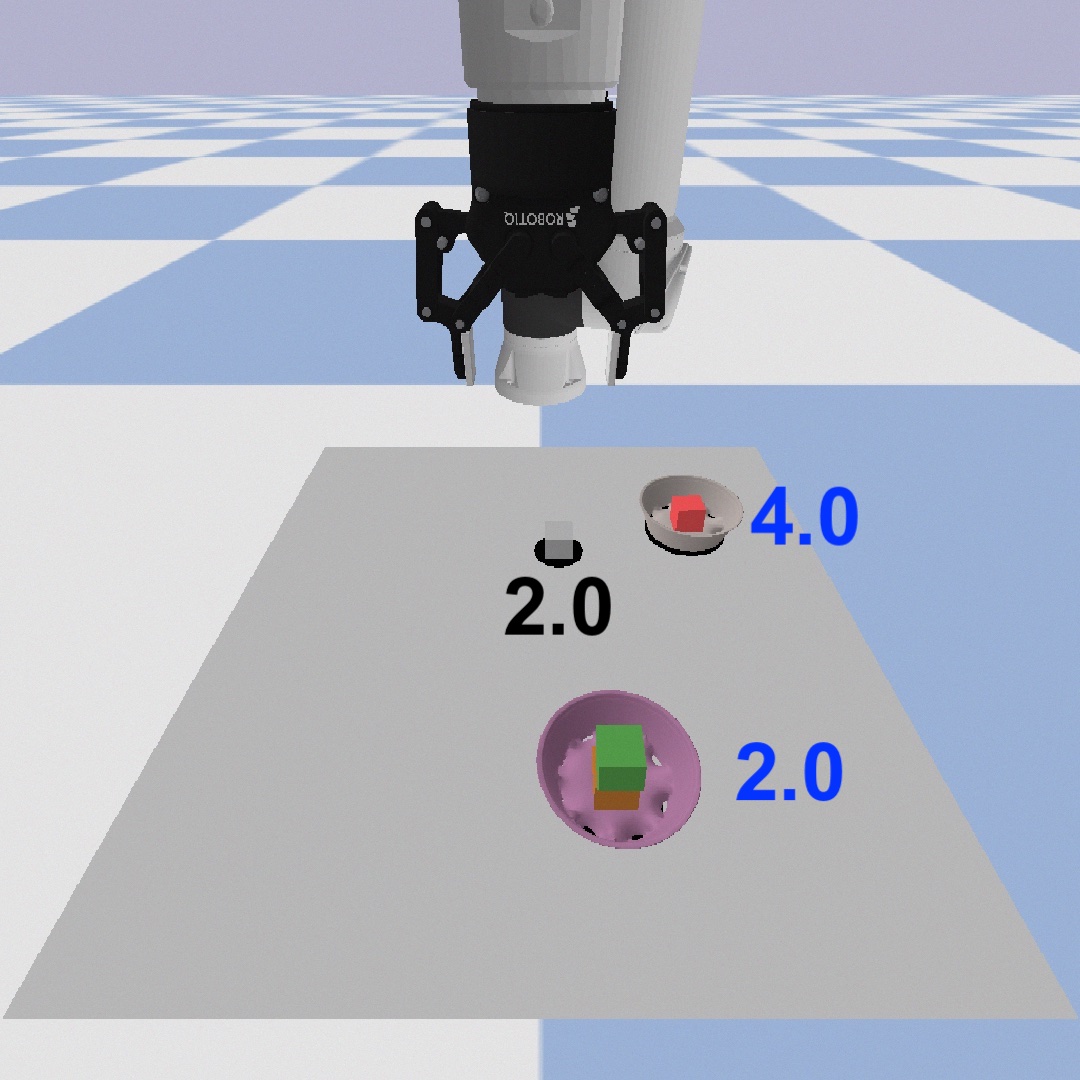}};
            \node [empty, right=0 of after-cap, label=below:{\footnotesize \acronym (ours)}] (after-ours) {\includegraphics[width=0.32\linewidth]{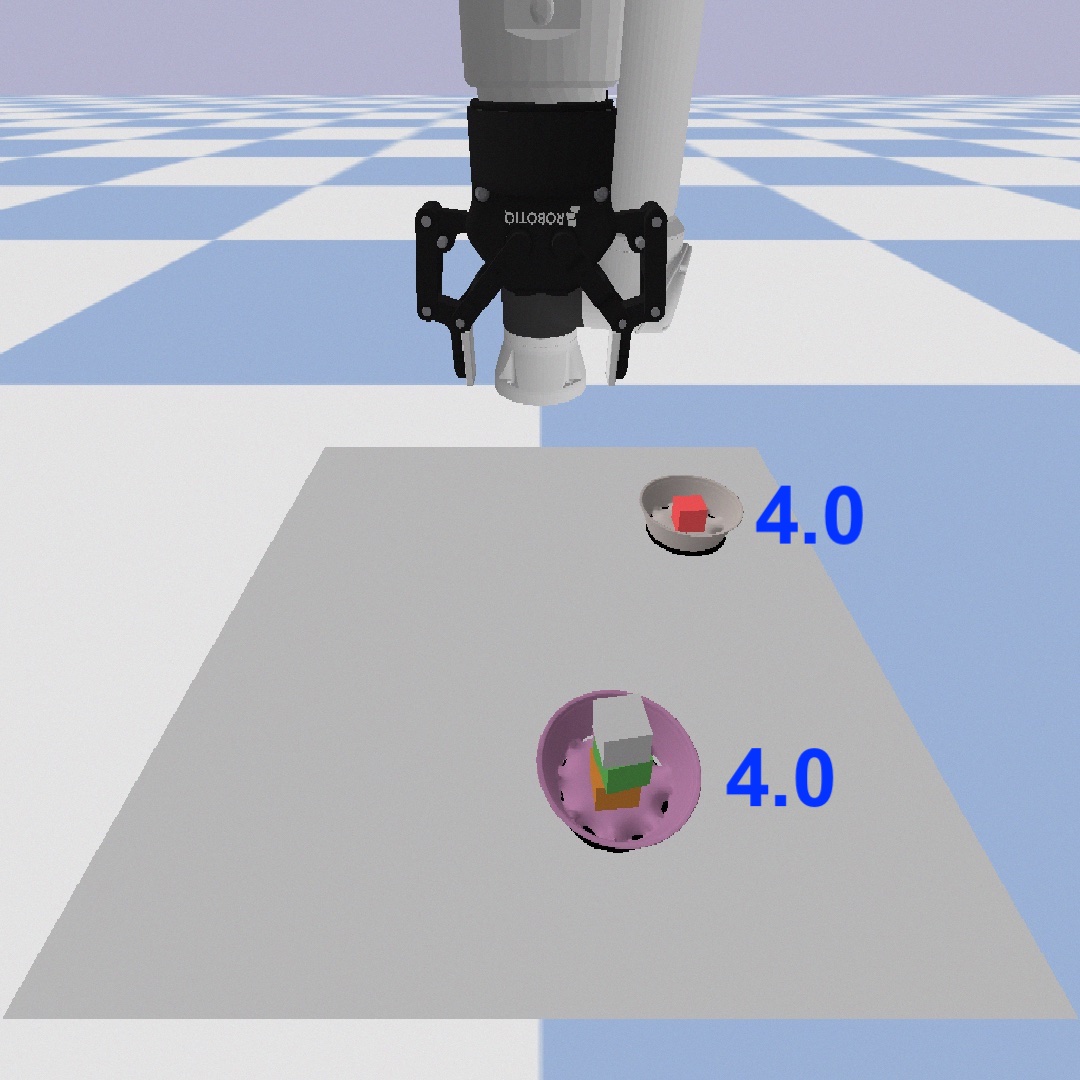}};
            \node[empty, above=0 and 0 of before.north west,anchor=south west] (speak) {\includegraphics[height=5mm]{figures/speak-icon}};
            \node[utterance, right= 1pt and 0 of speak.south east, anchor=south west, rounded corners] (utterance) {put blocks in the purple bowl so that their total weight becomes identical to what is in the gray bowl};
            \node [circle,draw=black, fill=white, inner sep=1pt, minimum size=5pt,  above right=5pt of after-ours.south west, anchor=south west] (check) {\greencheck};
            \node [circle,draw=black, fill=white, inner sep=1pt, minimum size=5pt, above right=5pt of after-cap.south west, anchor=south west] (check) {\redx};
        \end{tikzpicture}
    \end{subfigure} \hfil
    \begin{subfigure}{0.485\textwidth}
        \centering
        \begin{tikzpicture}[utterance/.style={rectangle, rounded corners, draw=black, font=\scriptsize, minimum width=5.75cm, minimum height=0.5cm, text width = 5.75cm, fill=black!10, outer sep=0pt},
            empty/.style={outer sep=1pt, inner sep=0pt}]
            \node [empty, label=below:{\footnotesize Initial State}] (before) at (0,0) {\includegraphics[width=0.32\linewidth]{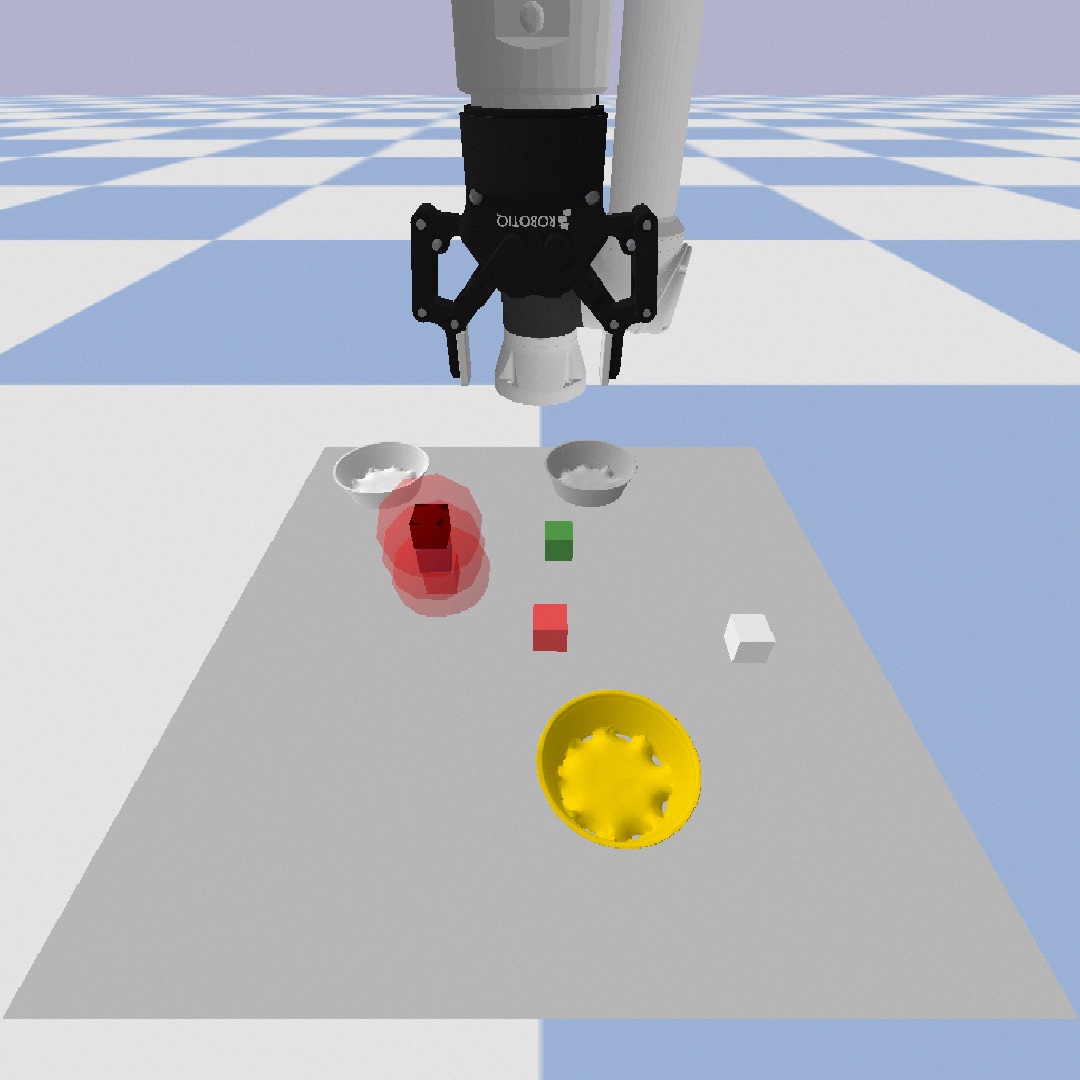}};
            \node [empty, right=0 of before, label=below:{\footnotesize \vanillallm}] (after-cap) {\includegraphics[width=0.32\linewidth]{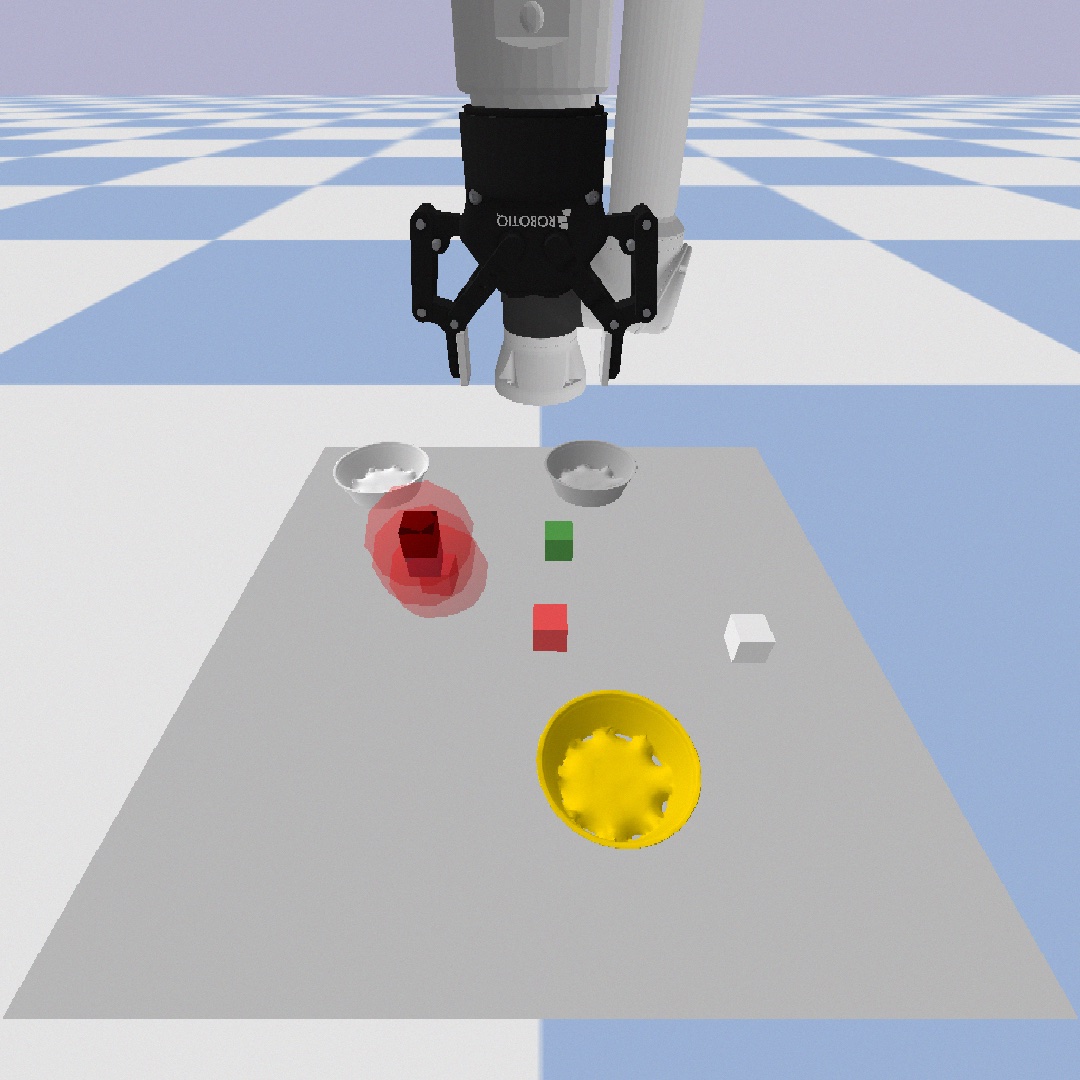}};
            \node [empty, right=0 of after-cap, label=below:{\footnotesize \acronym (ours)}] (after-ours) {\includegraphics[width=0.32\linewidth]{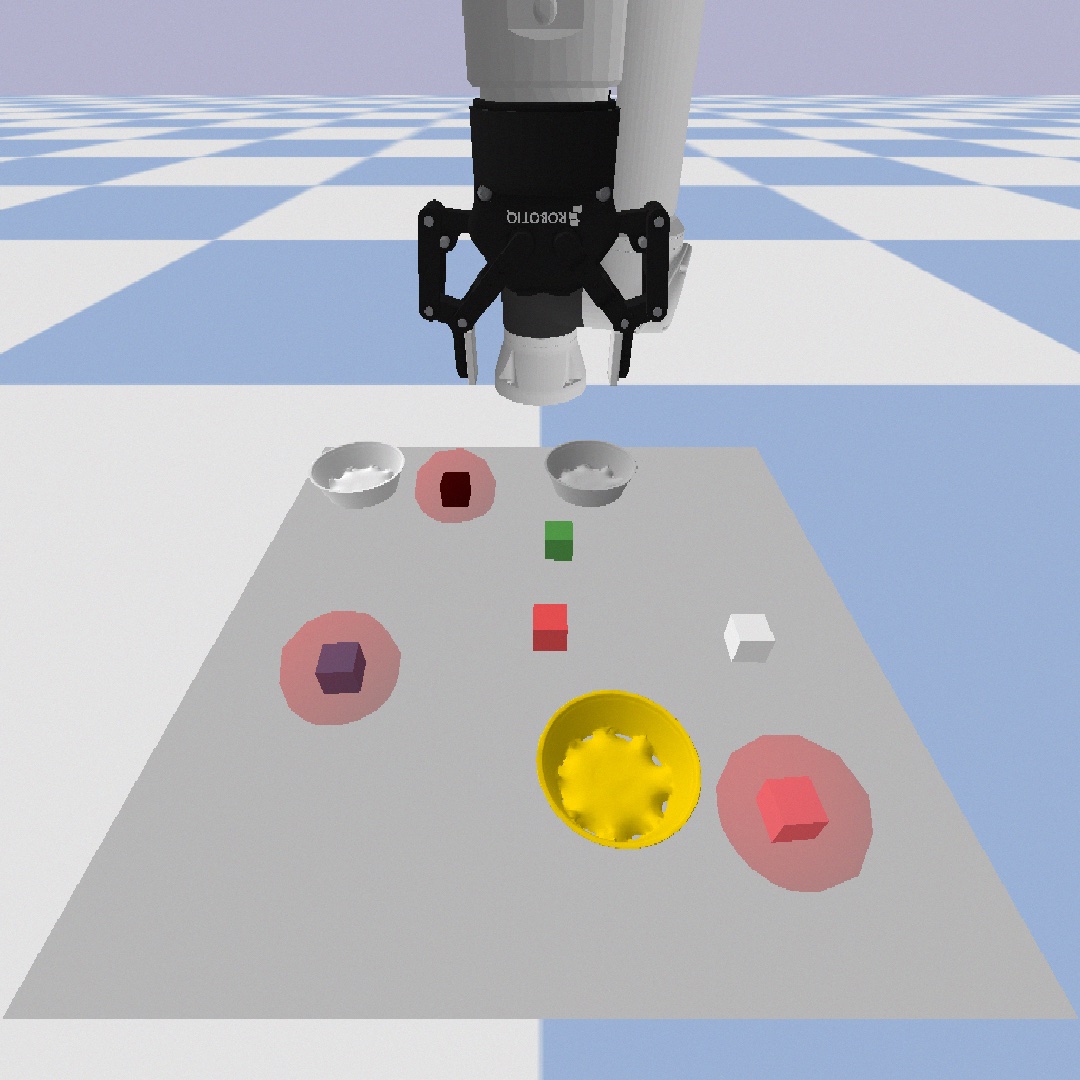}};
            \node[empty, above=0 and 0 of before.north west,anchor=south west] (speak) {\includegraphics[height=5mm]{figures/speak-icon}};
            \node[utterance, right= 1pt and 0 of speak.south east, anchor=south west, rounded corners] (utterance) {put all the dirty blocks on the table};
            \node [circle,draw=black, fill=white, inner sep=1pt, minimum size=5pt,  above right=5pt of after-ours.south west, anchor=south west] (check) {\greencheck};
            \node [circle,draw=black, fill=white, inner sep=1pt, minimum size=5pt, above right=5pt of after-cap.south west, anchor=south west] (check) {\redx};
        \end{tikzpicture}
    \end{subfigure}
    \caption{Examples of simulations that show the result of executing different natural language instructions using \vanillallm and our state-maintaining \acronym algorithm.}\label{fig:example-instructions-simulated}
\end{figure}
\begin{codefloat}
\begin{minipage}[t]{\linewidth}
\begin{minipage}[t]{\linewidth}
\begin{lstlisting}[caption={World-state reader. The text \sethlcolor{aigreen}\hl{highlighted in green} represents the part that the model is expected to generate.},label={lst:wm-reader},firstnumber=auto]
# @\label{line:init-metal-state}@state = {
#     "objects": ["cyan block", "yellow block", "brown block", "purple block", "blue block", "green bowl", "red bowl", "disinfector"],
#     "relations": [],
#     "disinfector": {"contains": []},
#     "cyan block": {"is": ["dirty"]},
#     "yellow block": {"is": ["clean"]},
#     "brown block": {"is": ["clean"]},
#     "purple block": {"is": ["dirty"]},
#     "blue block": {"is": ["clean"]},
#     "green bowl": {},
#     "red bowl": {}
# }@\label{line:end-metal-state}@
@
\label{line:metal-query}\textcolor{commentgray}{\# query: Put the cyan block on the yellow block}
@
@\sethlcolor{aigreen}\hl{put\_first\_on\_second("cyan block", "yellow block")}@
@\sethlcolor{aigreen}\hl{update\_wm("Put the cyan block on the yellow block")}@
\end{lstlisting}
\end{minipage}%
\hfill
\begin{minipage}[t]{\linewidth}
\begin{lstlisting}[caption={World-state writer. The text \textcolor{blue}{rendered in blue} highlights the updated part of the state.},label={lst:wm-writer},firstnumber=auto]
# state = {
#     "objects": ["cyan block", "yellow block", "brown block", "purple block", "blue block", "green bowl", "red bowl", "disinfector"],
#     "relations": [],
#     "disinfector": {"contains": []},
#     "cyan block": {"is": ["dirty"]},
#     "yellow block": {"is": ["clean"]},
#     "brown block": {"is": ["clean"]},
#     "purple block": {"is": ["dirty"]},
#     "blue block": {"is": ["clean"]},
#     "green bowl": {},
#     "red bowl": {}
# }
@\textcolor{commentgray}{\# query: Put the cyan block on the yellow block.}@
@\sethlcolor{aigreen}\hl{\#~state = \{}@
@\sethlcolor{aigreen}\hl{\#~~~~~"objects": ["cyan block", "yellow block", "brown block", "purple block", "blue block", "green bowl", "red bowl", "disinfector"],}@
@\textcolor{blue}{\sethlcolor{aigreen}{\sethlcolor{aigreen}\hl{\#~~~~~"relations": [["cyan block is on yellow block"]],}}}@
@\sethlcolor{aigreen}\hl{\#~~~~~"disinfector": \{"contains": []\},}@
@\sethlcolor{aigreen}\hl{\#~~~~~"cyan block": \{"is": ["dirty"]\},}@
@\textcolor{blue}{\sethlcolor{aigreen}\hl{\#~~~~~"yellow block": \{"is": ["dirty"]\},}}@
@\sethlcolor{aigreen}\hl{\#~~~~~"brown block": \{"is": ["clean"]\},}@
@\sethlcolor{aigreen}\hl{\#~~~~~"purple block": \{"is": ["dirty"]\},}@
@\sethlcolor{aigreen}\hl{\#~~~~~"blue block": \{"is": ["clean"]\},}@
@\sethlcolor{aigreen}\hl{\#~~~~~"green bowl": \{\},}@
@\sethlcolor{aigreen}\hl{\#~~~~~"red bowl": \{\},}@
@\sethlcolor{aigreen}\hl{\# \}}@
\end{lstlisting}
\end{minipage}
\hfill
\end{minipage}
\end{codefloat}
As exemplified in \cref{sec:motivation}, the key to our approach is to allow the LLM to describe the next state while responding to each user query. The motivating example is simple in that the next state \textit{is} the response. Instead, we now consider more general scenarios where there is a significant burden on the LLM to track the state updates as well as generate responses. (Fig.~\ref{fig:example-instructions-simulated}). %
For the general cases,
we propose to \emph{split} the burden across multiple different prompted LLMs. 
Precisely, we maintain a separate prompt that includes instructions and demonstrations for each subtask (i.e., state-tracking or query-responding) and then use the prompt to elicit an LLM to perform the particular subtask. 
As we discuss shortly, our framework includes a \textbf{world-state reader} that responds to the user query and a \textbf{world-state writer} that is responsible for updating the state representation.
Our framework (Fig.~\ref{fig:diagram}) does not pose any fundamental limitation on which domain it can be applied to.
Our approach can be regarded as a model-based extension of Code-as-Policies (CaP) in the sense that it keeps the core capabilities of CaP (e.g., hierarchical code generation) and incorporates a means to explicitly maintain an estimated world state.

It is useful to consider example prompts to understand the operation of the reader and writer models.
Prompt~\ref{lst:wm-reader} is an example of the input passed to the world-state reader. 
Initially, we initialize a JSON-formatted state with a reference to object-oriented principles.
Given a user query ``Put the cyan block on the yellow block'' (Line \ref{line:metal-query}) and the current state representation (Lines \ref{line:init-metal-state}--\ref{line:end-metal-state}),
the world-state reader should generate the code that responds to the query, taking into account the current state. The expected code to be generated is \sethlcolor{aigreen}\hl{highlighted in green}.
After generating the code, our model executes it to complete the query. When the state needs to be updated, the generated code will contain an \texttt{update\_wm} function that triggers the world-state writer with the query specified in its argument.
In Prompt~\ref{lst:wm-writer}, we show the corresponding example for the world-state writer. Similar to the reader, we prepend the current state representation before the user query and the model generates the updated state representation (\hl{highlighted in green}).
 Whenever the writer updates the state representation, we store it in external memory and refer to it as the current state.

\section{Experiments}\label{sec:experiments}

\begin{figure}[!t]
    \centering
    \begin{subfigure}{0.15\textwidth}
        \centering
        \includegraphics[width=\linewidth]{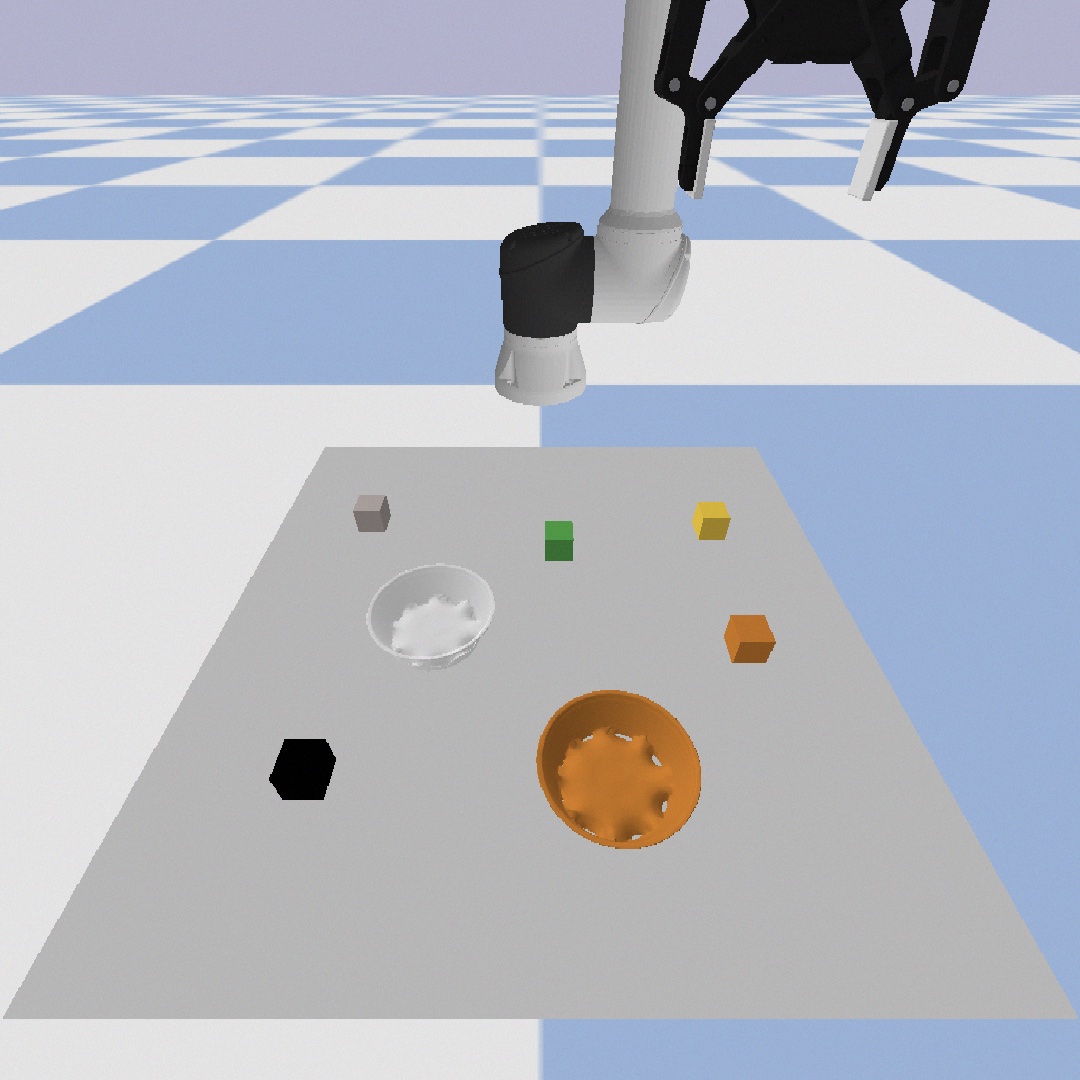}
        \caption{Pick-and-place}\label{fig:pick-and-place}
    \end{subfigure} \hfil
    \begin{subfigure}{0.15\textwidth}
        \centering
        \includegraphics[width=\linewidth]{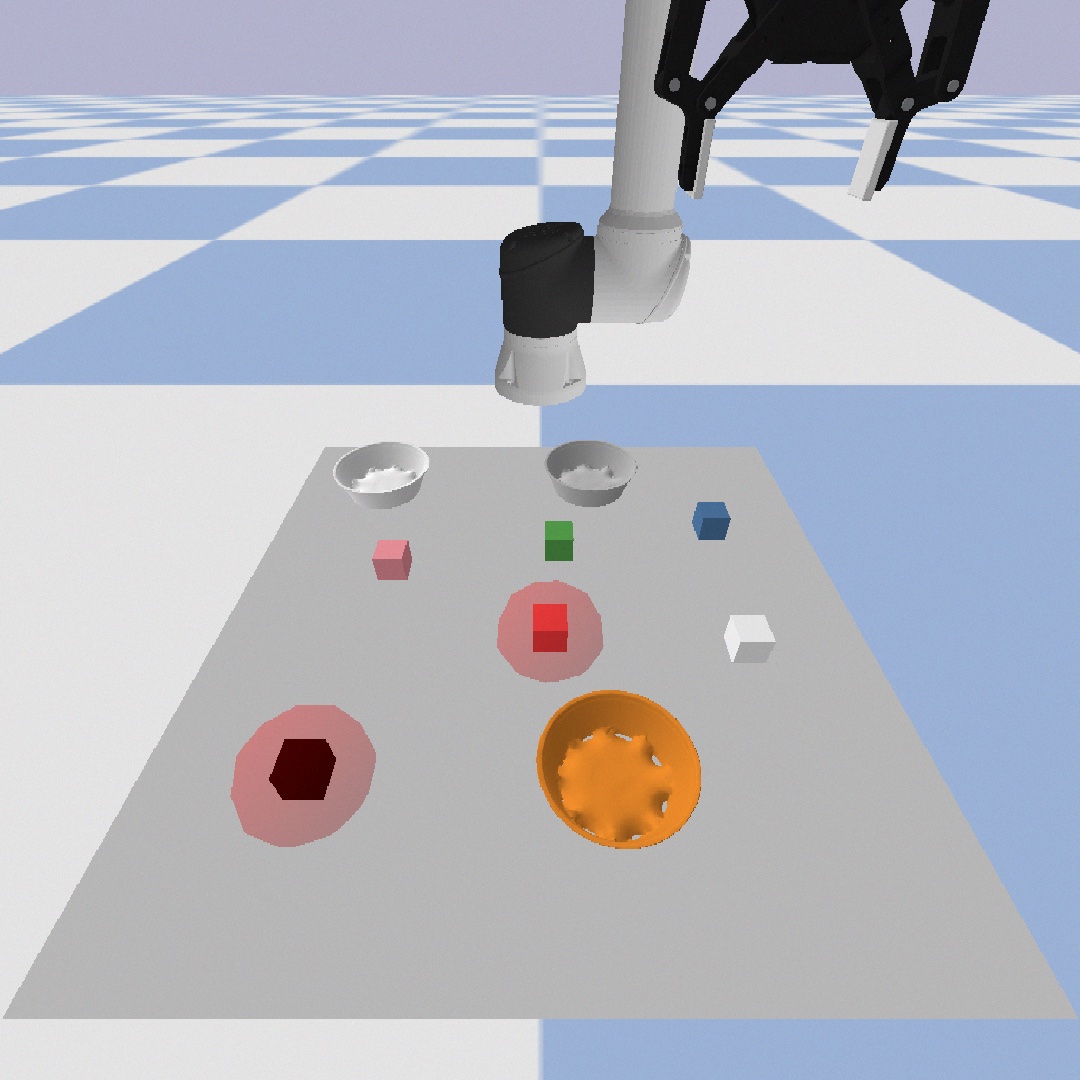}
        \caption{Disinfection}\label{fig:disinfection}
    \end{subfigure} \hfil
    \begin{subfigure}{0.15\textwidth}
        \centering
        \includegraphics[width=\linewidth]{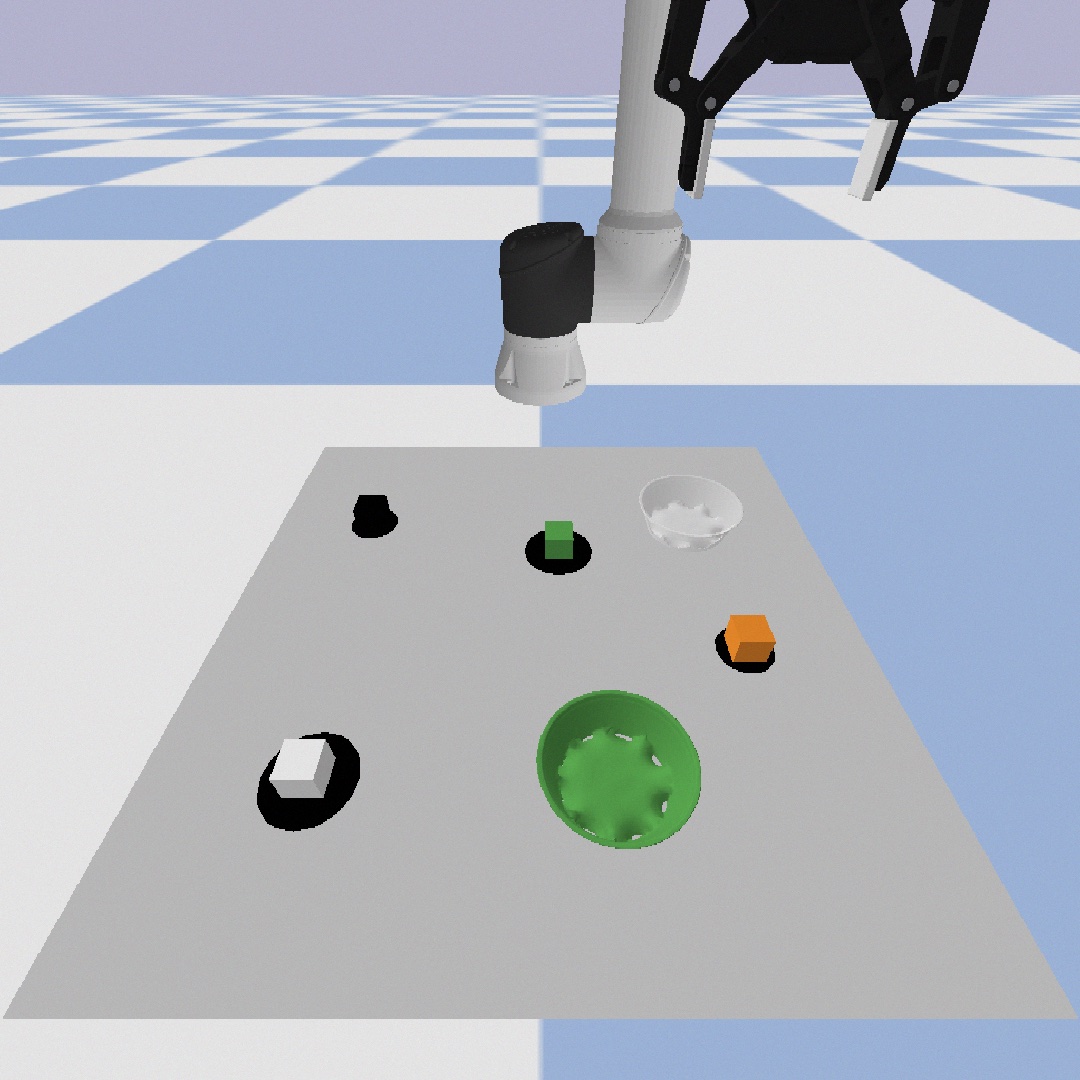}
        \caption{Weight reasoning}\label{fig:weight-reasoning}
    \end{subfigure}
    \caption{The simulated domains we consider include \subref{fig:pick-and-place} Pick-and-Place; \subref{fig:disinfection} Block Disinfection, where the translucent sphere around a block represents its dirtiness (this is not visible to the robot); and \subref{fig:weight-reasoning} Relative Weight Reasoning, where the radius of the disk under each block indicates its weight (this is not visible to the robot).}\label{fig:bp-assisted}
\end{figure}

We evaluate the capabilities of \acronym alongside state-of-the-art LLM models on three tabletop manipulation domains (Fig.~\ref{fig:bp-assisted}): pick-and-place, block disinfection, and relative weight reasoning. For each domain, we design in-context examples and consider $20$ evaluation episodes each of which consists of $5$--$16$ consecutive steps of user queries. Every episode contains at least one query that requires reasoning over the interaction history (i.e., requires ``memory'' across steps), which makes the task significantly  challenging.%

\subsection{Simulated Tabletop Manipulation Domains}
The \textbf{Pick-and-Place} domain involves scenarios that require a robot arm to sequentially pick up and place a block onto another block, bowl, or the table. The model needs to remember and reason over the block locations. The example user queries are ``Put the green block in the red bowl.'', ``What is the color of the block under the pink block?'', and ``How many blocks are in the green bowl?''.%

In the \textbf{Block Disinfection} domain, we consider a scenario in which a block can be either \textit{dirty} or \textit{clean}, the state of which is not observable by the robot. When a clean block touches a dirty block (e.g., as a result of stacking one block on another), the clean block becomes dirty. There is a \textit{disinfector} on the table that cleans any block placed inside it. This scenario emulates a clean-up task in which you might ask a robot to put dirty dishes in a dishwasher or dirty clothes in a washing machine. The user query contains pick-and-place commands similar to those in the pick-and-place domain as well as textual utterances that require reasoning over which blocks are clean and dirty, such as ``Put all the clean blocks in the green bowl.'' This domain presents a particular challenge as the model must track the cleanliness of each block and accurately capture the state mutations that happen when a dirty block comes into contact with another clean block.

\textbf{Relative Weight Reasoning} involves memorizing and reasoning over the relative weights of the blocks. User queries provide information about the weight of blocks (e.g., ``The red block is twice the weight of the bronze block.''), which are followed by queries that require reasoning over the weights (e.g., ``Put blocks in the purple bowl so that their total weight becomes identical to what is in the gray bowl.'').

We compare our proposed approach, \acronym, to two strong competing methods: Code-as-Policies~\cite{Liang2022CodeAP} (CaP) and CaP with Chain-of-Thought prompting~\cite{DBLP:conf/nips/Wei0SBIXCLZ22} (CaP+CoT).%
CaP generates code for the current question at each step based on the past actions, but it does not maintain a state.
Following the CoT framework, \textit{at every step}, CaP+CoT deduces the intermediate states based on an \textit{initial state} and past actions, which are considered as its thoughts, to generate the current code. 
But it leads to redundant reasoning and increases the length of the prompt, which may then exceed the LLM's context window size limitations.
Furthermore, longer reasoning also demands longer, more intricate demo example prompts, contributing to increased developer effort.
We ensure that the demonstrations (i.e., in-context examples) given to each of the models are equivalent. Namely, we use the same sequence of user queries and code snippets, except for necessary differences due to their designs such as state representation.

\begin{figure}[!t]
    \centering
        \begin{tikzpicture}[utterance/.style={rectangle, rounded corners, draw=black, font=\scriptsize, minimum width=5.75cm, minimum height=0.5cm, text width = 5.75cm, fill=black!10, outer sep=0pt},
        response/.style={rectangle, draw=black, font=\scriptsize, text width=4.05cm, outer sep=0pt}, empty/.style={outer sep=0pt, inner sep=0pt}]
            \node [empty] (before) at (0,0) {\includegraphics[width=0.24\linewidth]{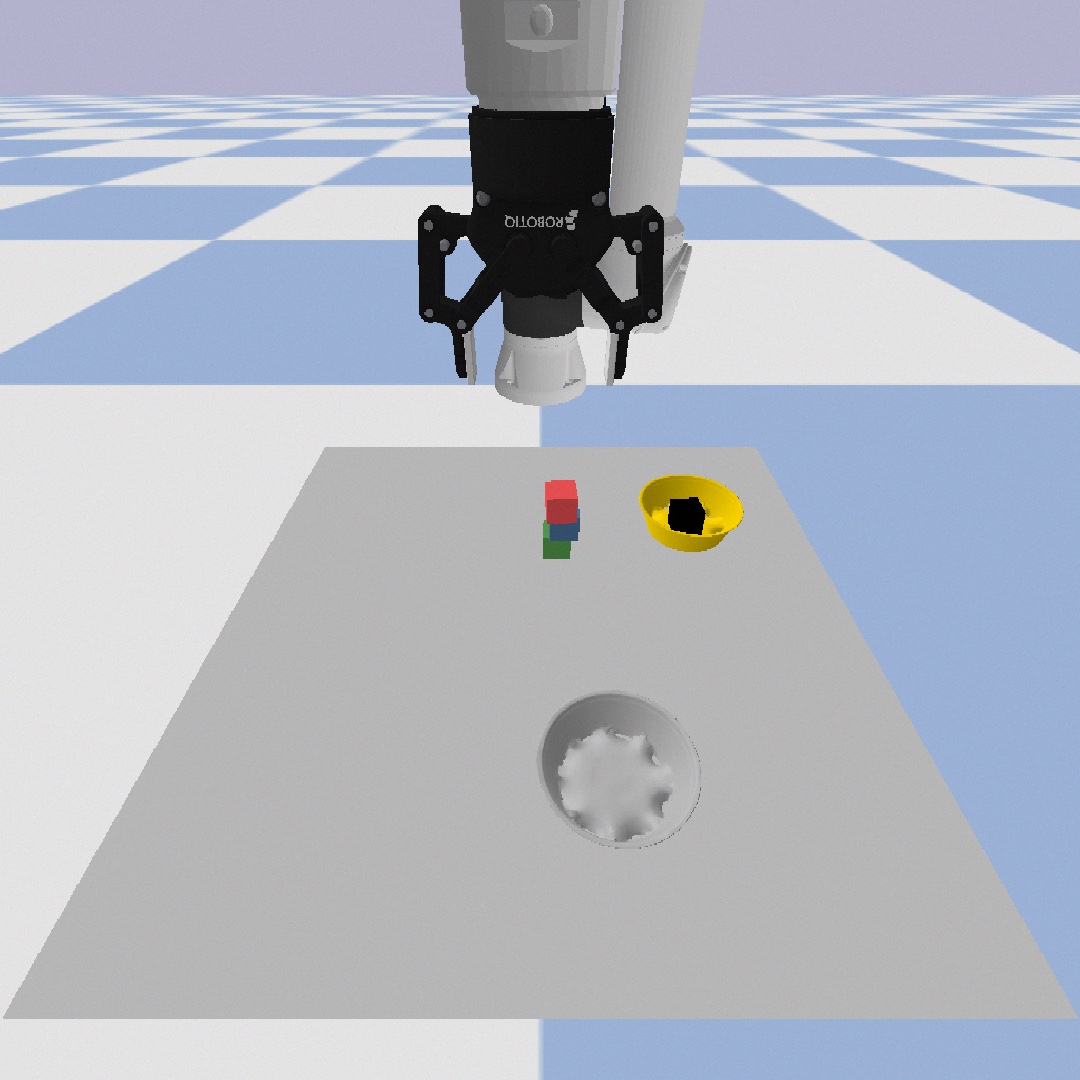}};
            \node[empty, above=0 and 0 of before.north west,anchor=south west] (speak) {\includegraphics[height=5mm]{figures/speak-icon}};
            \node[utterance, right= 1pt and 0 of speak.south east, anchor=south west, rounded corners] (utterance) {\textls[-20]{What is the color of the block right above the blue block?}};
            \path   let \p1 = ($(before.north east)-(utterance.north east)$),
                        \p2 = ($(before.north east)-(before.east)$),
                        \n1 = {veclen(\x1,\y1)},
                        \n2 = {veclen(\x2,\y2)}, in
                    node[response, minimum width=\n1, minimum height=\n2, right=1pt of before.north east,anchor=north west] (input) {\scriptsize \textbf{\vanillallm}: \textit{fails to generate anything}};
            \path   let \p1 = ($(before.north east)-(utterance.north east)$),
                        \p2 = ($(before.north east)-(before.east)$),
                        \n1 = {veclen(\x1,\y1)},
                        \n2 = {veclen(\x2,\y2)}, in
                    node[response, minimum width=\n1, minimum height=\n2, right=1pt of before.east,anchor=north west] (input) {\scriptsize \textbf{\acronym (ours)}: ``red''};
        \end{tikzpicture}
        \begin{tikzpicture}[utterance/.style={rectangle, rounded corners, draw=black, font=\scriptsize, minimum width=5.75cm, minimum height=0.5cm, text width = 5.75cm, fill=black!10, outer sep=0pt},
            response/.style={rectangle, draw=black, font=\scriptsize, outer sep=0pt}, empty/.style={outer sep=0pt, inner sep=0pt}]
                \node [empty] (before) at (0,0) {\includegraphics[width=0.24\linewidth]{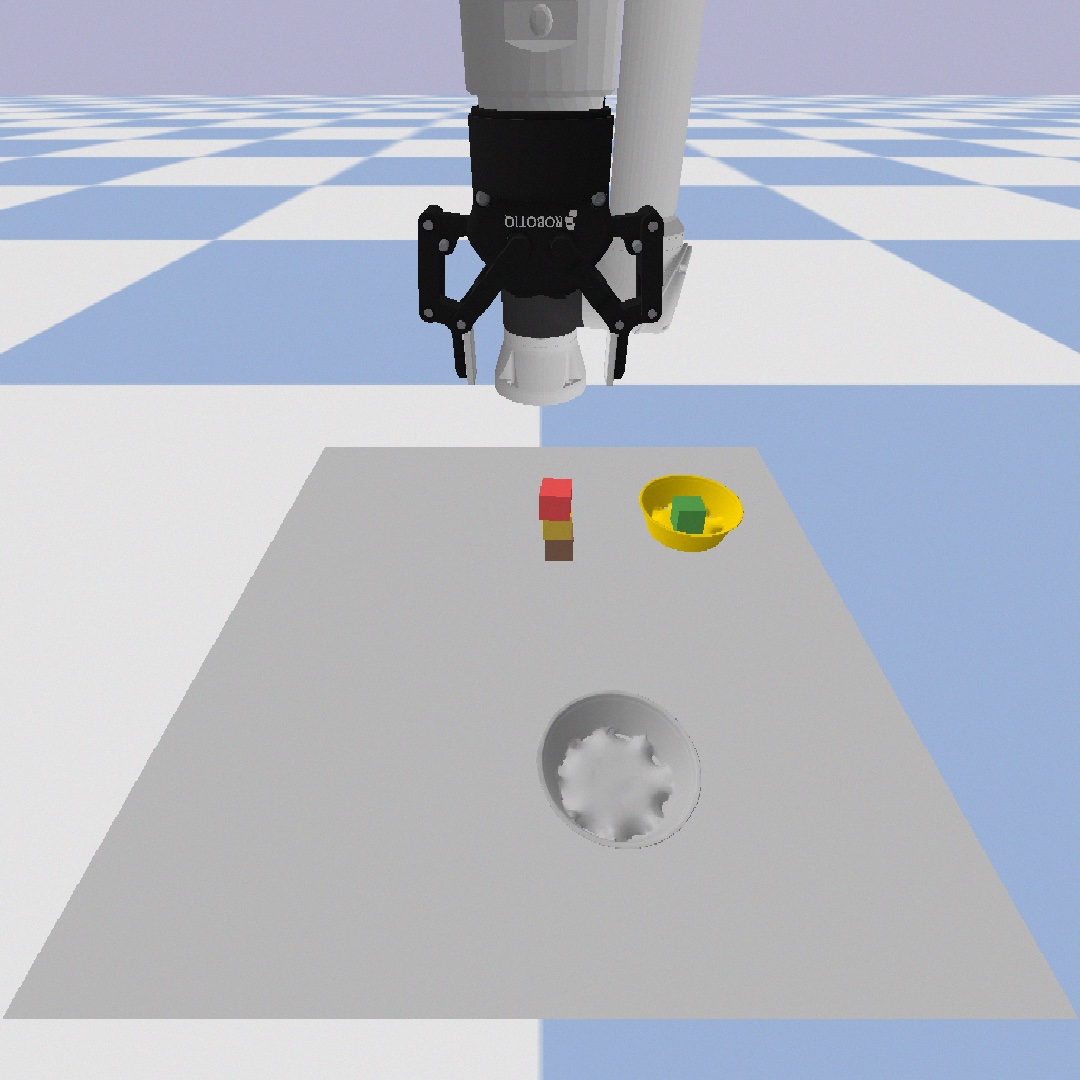}};
                \node[empty, above=0 and 0 of before.north west,anchor=south west] (speak) {\includegraphics[height=5mm]{figures/speak-icon}};
                \node[utterance, right= 1pt and 0 of speak.south east, anchor=south west, rounded corners] (utterance) {How many blocks are not in the bowls?};
                \path   let \p1 = ($(before.north east)-(utterance.north east)$),
                            \p2 = ($(before.north east)-(before.east)$),
                            \n1 = {veclen(\x1,\y1)},
                            \n2 = {veclen(\x2,\y2)}, in
                        node[response, minimum width=\n1, text width=4.0cm, minimum height=\n2, right=1pt of before.north east,anchor=north west] (input) {\textbf{\vanillallm}: ``There are two blocks not in the bowls: brown block and yellow block.''};
                \path   let \p1 = ($(before.north east)-(utterance.north east)$),
                            \p2 = ($(before.north east)-(before.east)$),
                            \n1 = {veclen(\x1,\y1)},
                            \n2 = {veclen(\x2,\y2)}, in
                        node[response, minimum width=\n1, text width=4.0cm, minimum height=\n2, right=1pt of before.east,anchor=north west] (input) {\scriptsize \textbf{\acronym (ours)}: ``three blocks''};
            \end{tikzpicture}
    \caption{Examples that show the result of querying LLMs with and without maintained state. In the first scenario, CaP fails to produce an answer, while our \acronym model produces the correct response. In the second example, one block is not visible and CaP incorrectly identifies two blocks as not being a bowl. By maintaining a persistent world state, our method is aware of the third block and correctly answers the query.}\label{fig:example-queries}
\end{figure}

\begin{table}[!th]
    \centering
    \caption{Episode success rates and individual step success rates (in parentheses) for each sequential task. $^\dagger$indicates that the context limit was often exceeded.} \label{tb:main-result}
    \setlength{\tabcolsep}{3pt}
    \begin{tabularx}{1.0\linewidth}{lYYY}%
        \toprule
         & {Pick \& Place} & {Disinfection} & {Weight}\\
        \midrule
        CaP & 
        \nfrac{0.00}{(0.54)} &%
        \nfrac{0.00}{(0.68)}\hphantom{${^\dagger}$} & %
        \nfrac{0.00}{(0.84)}\\
       CaP+CoT &
        \nfrac{0.25}{(0.76)} & %
        \nfrac{0.00}{(0.20)}${^\dagger}$ & %
        \nfrac{0.30}{(0.88)}\\
       \acronym (ours) &
        \nfracb{0.50}{(0.88)} &%
        \nfracb{0.40}{(0.82)}\hphantom{${^\dagger}$} &%
        \nfracb{0.55}{(0.93)} \\
       \bottomrule
    \end{tabularx}
\end{table}
Table~\ref{tb:main-result} reports the episode success rates of each method along with the the success rate for individual steps. An episode is considered to be a failure if a model fails to respond to one of the user queries in the episode. While the CaP baseline correctly processes more than half of the individual steps in each domain, it fails to successfully complete any of the episodes. As we show later, CaP correctly processes most queries that do not require reasoning over previous steps (e.g.,``Put the red block on the blue block.''), but tends to generate incorrect (or no) code in response to queries that require reasoning over the history (e.g., ``Put all the dirty blocks in the pink bowl.'' and ``What is the color of the block under the purple block?'') (see \cref{fig:example-queries} (top)). CaP+CoT fares slightly better in the Pick-and-Place and Relative Weight Reasoning, but still fails in most episodes. In contrast, \acronym successfully handles the majority of these queries, demonstrating strong improvement over the others. It should be noted we explicitly chose queries that were challenging for LLM-based models, which partially accounts for why our model's scores show room for improvement.

\begin{table}[!th]
    \centering
    \caption{Success rates of \vanillallm (CaP) and \acronym for non-temporal and temporal queries.}%
    \label{tb:main-result-wo-individual-truncation}
    \setlength{\tabcolsep}{3pt}
    {\footnotesize%
    \begin{tabularx}{1.0\linewidth}{lYYYY}%
        \toprule
        & \multicolumn{2}{c}{Non-temporal} & \multicolumn{2}{c}{Temporal}\\
        \midrule
        & CaP & \acronym (ours)& CaP & \acronym (ours)\\
        \midrule
        Pick \& Place &
        \nfrac{1.00}{\hphantom{0}(62/62)\hphantom{0}} &
        \nfrac{1.00}{\hphantom{0}(68/68)\hphantom{0}} &%
        \nfrac{0.31}{(9/29)} &
        \nfracb{0.83}{(48/58)}\\
        Disinfection &
        \nfrac{0.99}{(148/149)} &
        \nfrac{0.98}{(164/168)} &
        \nfrac{0.05}{(1/20)} &
        \nfracb{0.65}{(15/23)}\\
        Weight &
        \nfrac{1.00}{(107/107)} &
        \nfrac{1.00}{(107/107)} &
        \nfrac{0.00}{(0 / 20)} &
        \nfracb{0.55}{(11/20)}\\
        \bottomrule
    \end{tabularx}}
\end{table}
In order to better understand the behavior of \acronym in comparison to Code-as-Policies, we analyze the success rates %
based on the type of user queries. Specifically, we categorize each query as either \emph{temporal} or \emph{non-temporal} depending on whether responding to the query necessitates temporal reasoning. We emphasize that contemporary methods, including the baselines that we consider, use non-temporal queries for evaluation. Table~\ref{tb:main-result-wo-individual-truncation} summarizes the resulting accuracy. 
The models often fail at different steps in an episode. We note that, when calculating accuracy we only consider the sequence of steps until %
the model fails to generate the correct code, which explains the mismatch in the denominators.

We see that both models achieve near-perfect performance on commands that do not require temporal reasoning. However, the performance of CaP noticeably decreases for tasks that require reasoning over the past. In contrast, \acronym achieves success rates of $83\%$ (vs.\ $31\%$ for CaP) on Pick-and-Place, $65\%$ (vs.\ $5\%$ for CaP) on Block Disinfection, and $55\%$ (vs. $0\%$ for CaP) on Relative Weight Reasoning.

\vspace{-1pt}

Although our method enjoys a better robustness than the baseline methods, it inherits some issues of large language models, which hinders its performance. For example, it hallucinates block conditions (e.g., clean or dirty) or locations when the cleanliness of the block is never explicitly described. Moreover, the model's reasoning strategy seems to predominantly treat weight as an abstract concept, e.g. light vs. heavy, rather than executing mathematical computations. This weakness is evident when asking the model to accumulate blocks in a bowl until their total weight surpasses that of another bowl, yet the model underfills the bowl. 
In the disinfection domain, our model struggles to comprehend ambiguous terms like ``other'' in queries such as ``the other blocks are clean.'' It can also wrongly infer from the training prompt that a block \emph{at the bottom} becomes dirty when a block is placed on top of it, irrespective of the latter's cleanliness.

\vspace{-1.5pt}

\subsection{Real Robot Experiments}

\vspace{-1.5pt}

\begin{figure}[!t]
    \centering
    \newlength{\mywidth}
    \setlength{\mywidth}{0.19\linewidth}%
    \begin{tikzpicture}[
        utterance/.style={rectangle, rounded corners, draw=black, font=\scriptsize, minimum width=4.25cm, minimum height=0.5cm, text width = 4.25cm, fill=black!10, outer sep=0pt},empty/.style={outer sep=1pt, inner sep=0pt}]
        \node[empty,rotate=90] at (0,0) (captext) {\bf\tiny Code-as-Policies};
        \node[empty, above=7pt and 5pt of captext.north east,anchor=south west] (speak) {\includegraphics[height=5mm]{figures/speak-icon}};
        \node[utterance, right= 1pt and 0 of speak.south east, anchor=south west, rounded corners] (utterance) {\textls[-20]{Put the black cup on the yellow block.\\ Put the yellow block on the Rubik's cube.}};
        \node[empty, right=0pt of captext.south] (cap0) {\includegraphics[width=\mywidth]{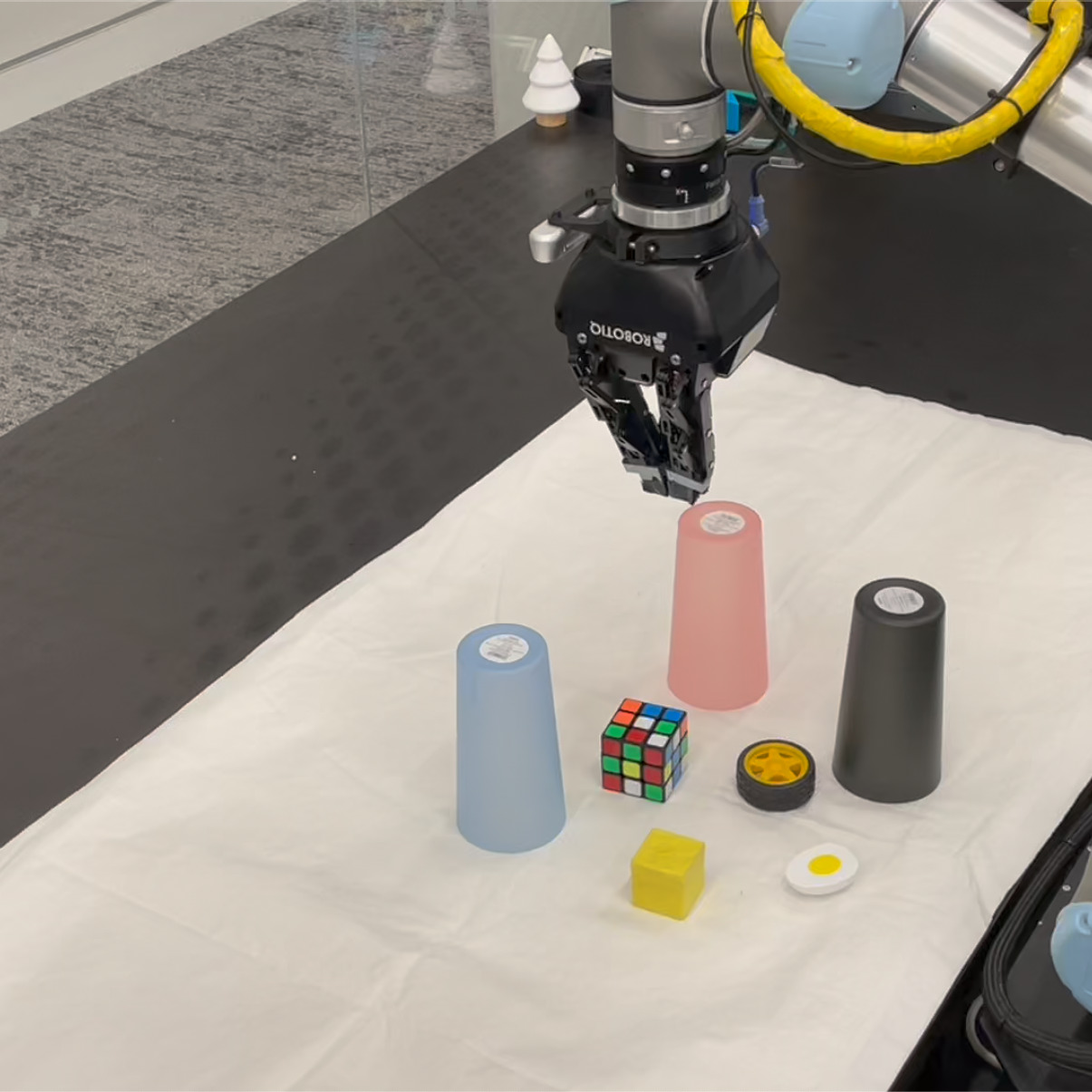}};
        \node[empty,right = -1pt of cap0, draw=green, ultra thick] (cap1) {\includegraphics[width=\mywidth]{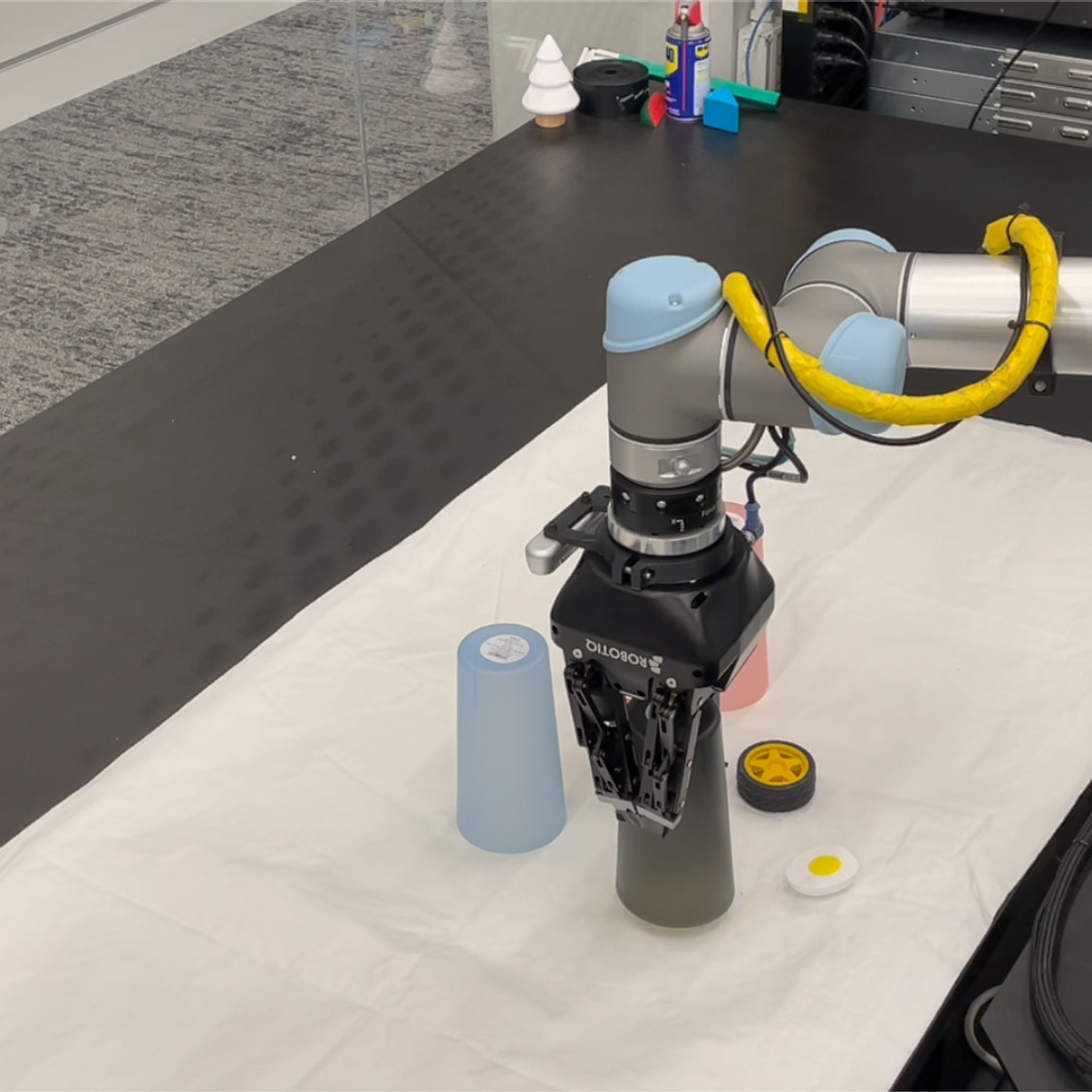}};
        \node[empty,right = -1pt of cap1] (cap2) {\includegraphics[width=\mywidth]{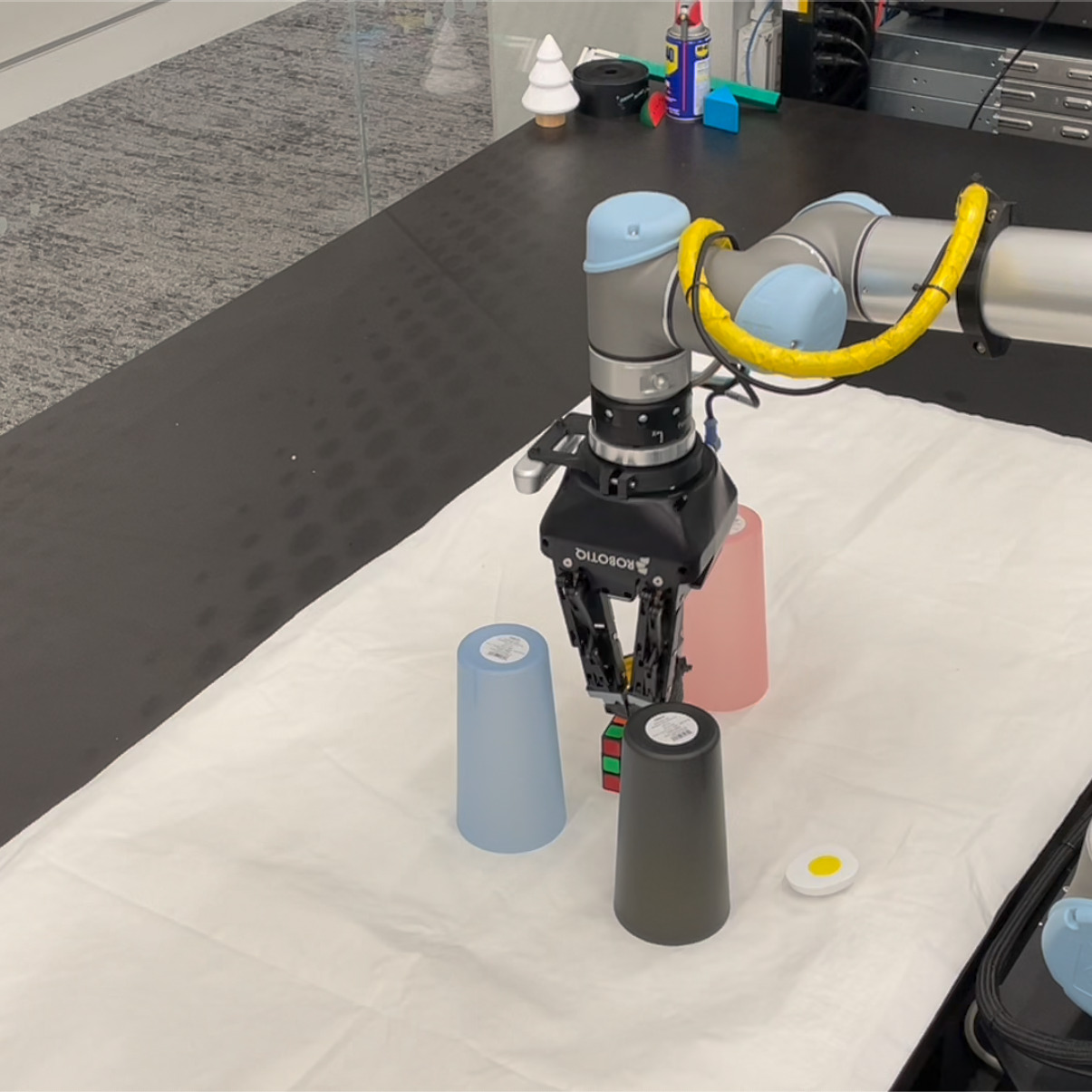}};
        \node[empty,right = -1pt of cap2,draw=red,ultra thick] (cap3) {\includegraphics[width=\mywidth]{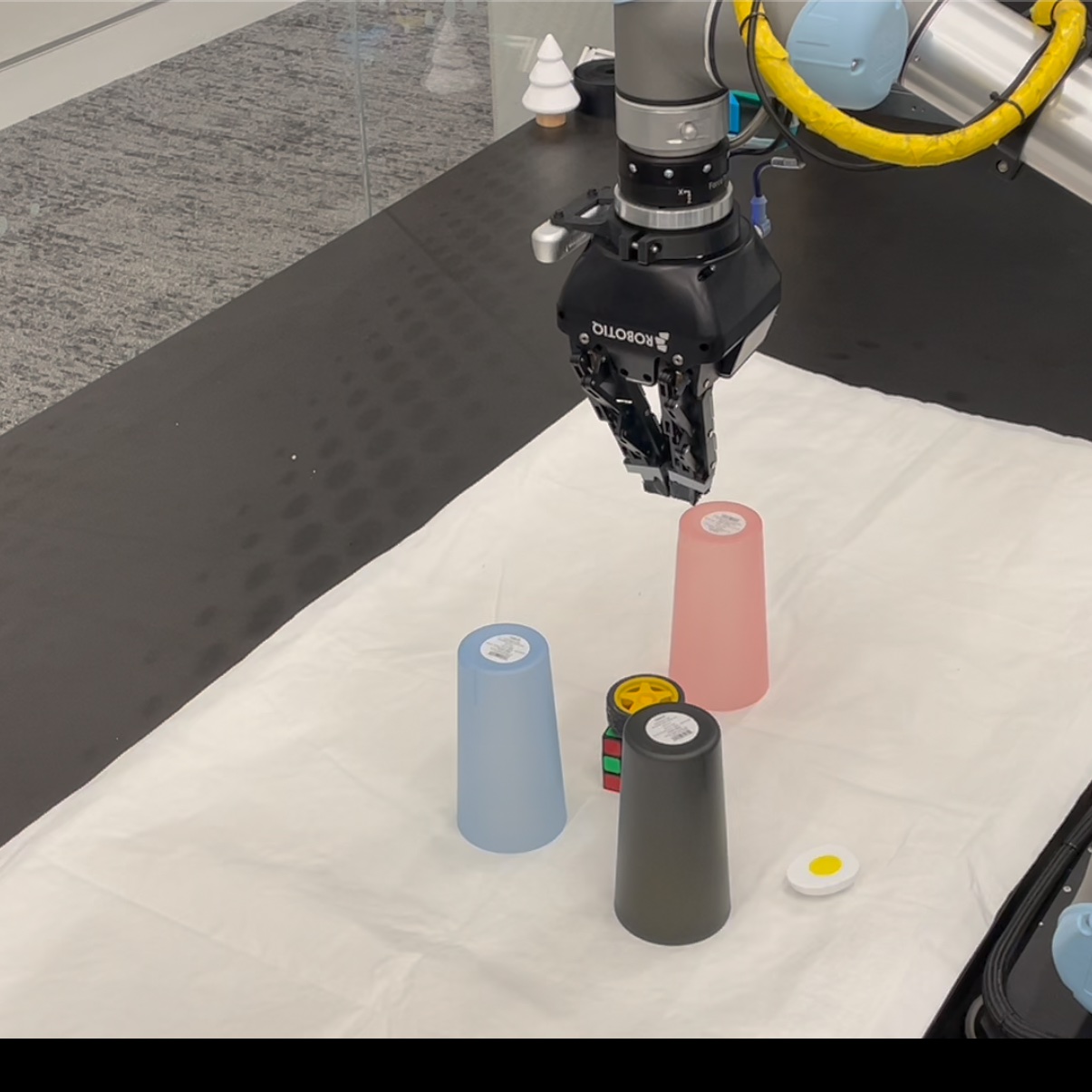}};
        \node [circle,draw=black, fill=white, inner sep=1pt, minimum size=5pt, above right=5pt of cap3.south west, anchor=south west] (check) {\redx};
        \node[empty,below = 1pt of cap0] (statler0) {\includegraphics[width=\mywidth]{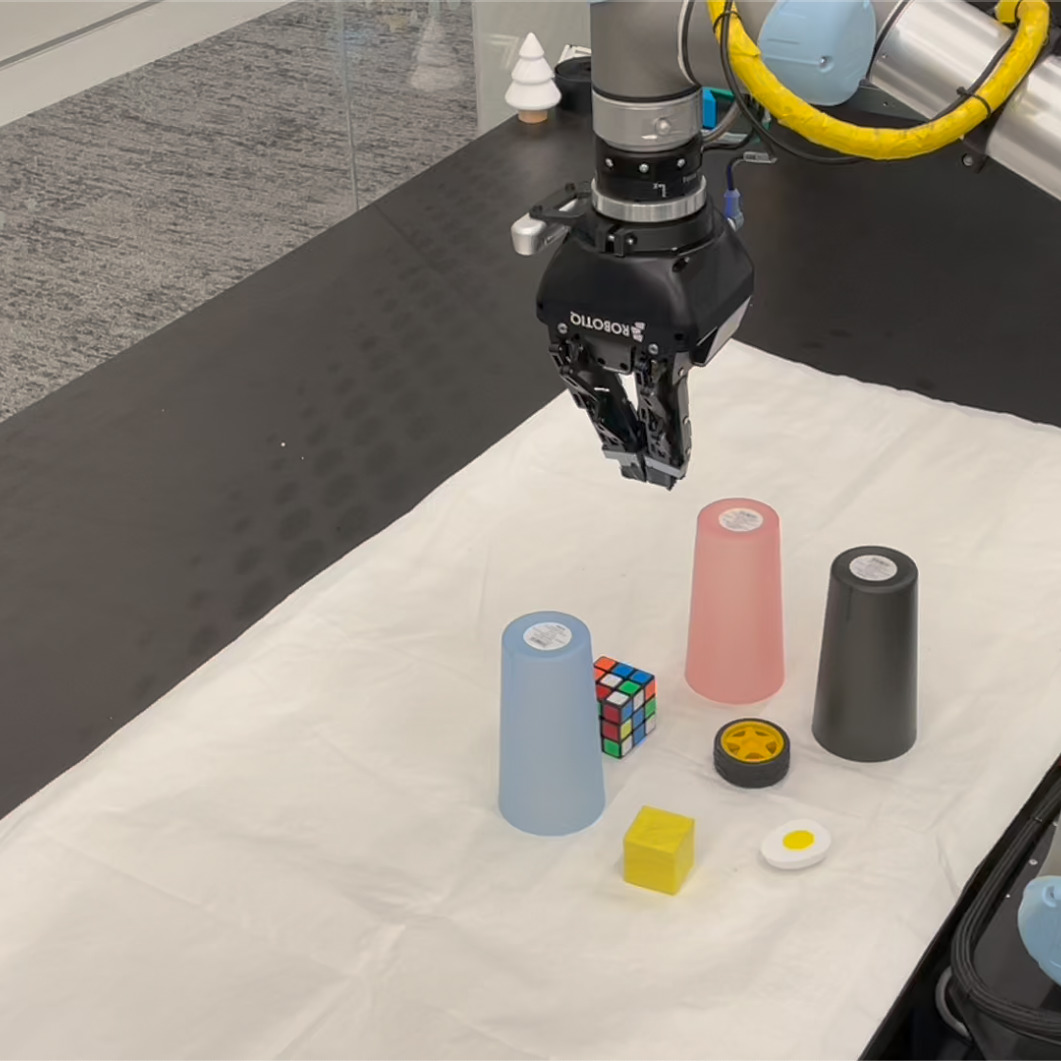}};%
        \node[empty,left = 0pt of statler0, rotate=90, anchor=south] () {\bf\tiny \acronym};
        \node[empty,right = -1pt of statler0, draw=green, ultra thick] (statler1) {\includegraphics[width=\mywidth]{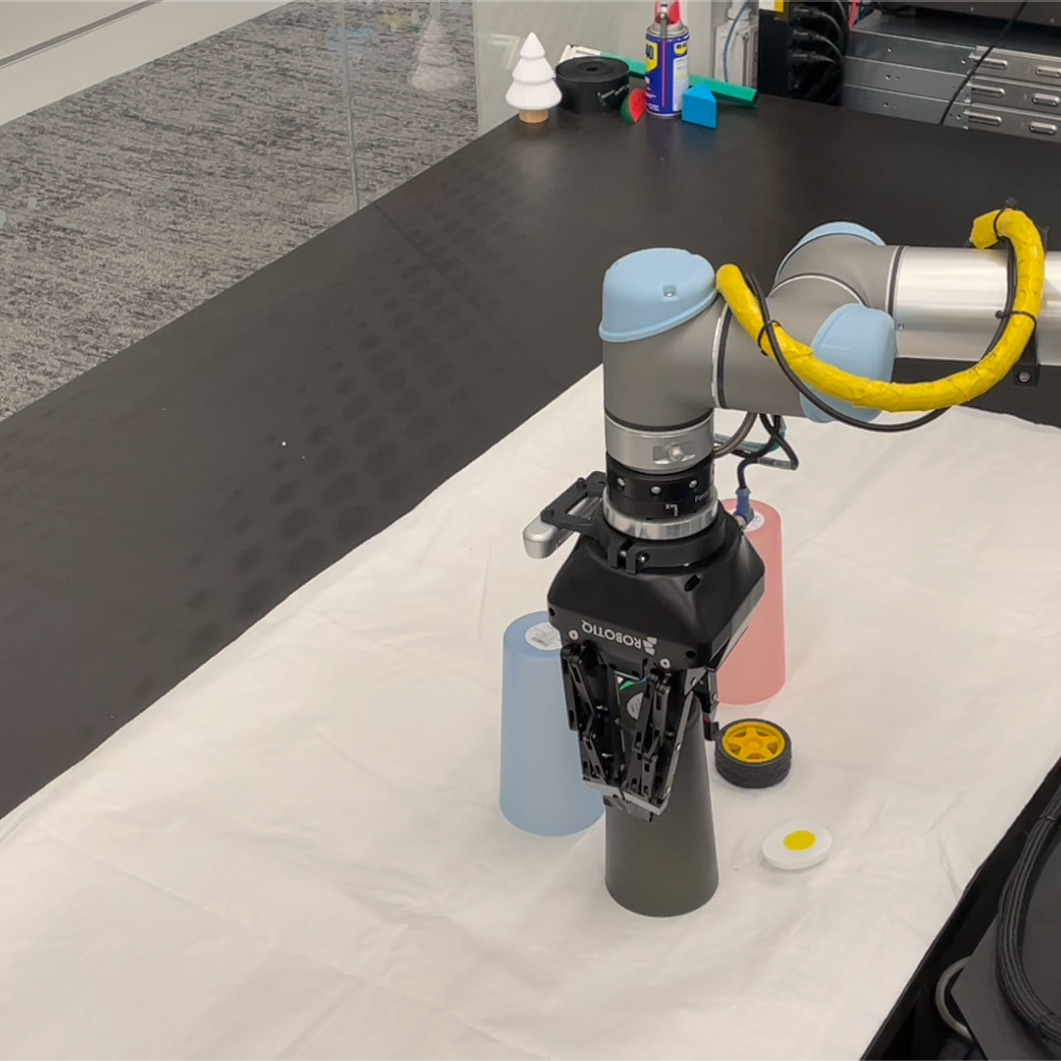}};%
        \node[empty,right = -1pt of statler1] (statler2) {\includegraphics[width=\mywidth]{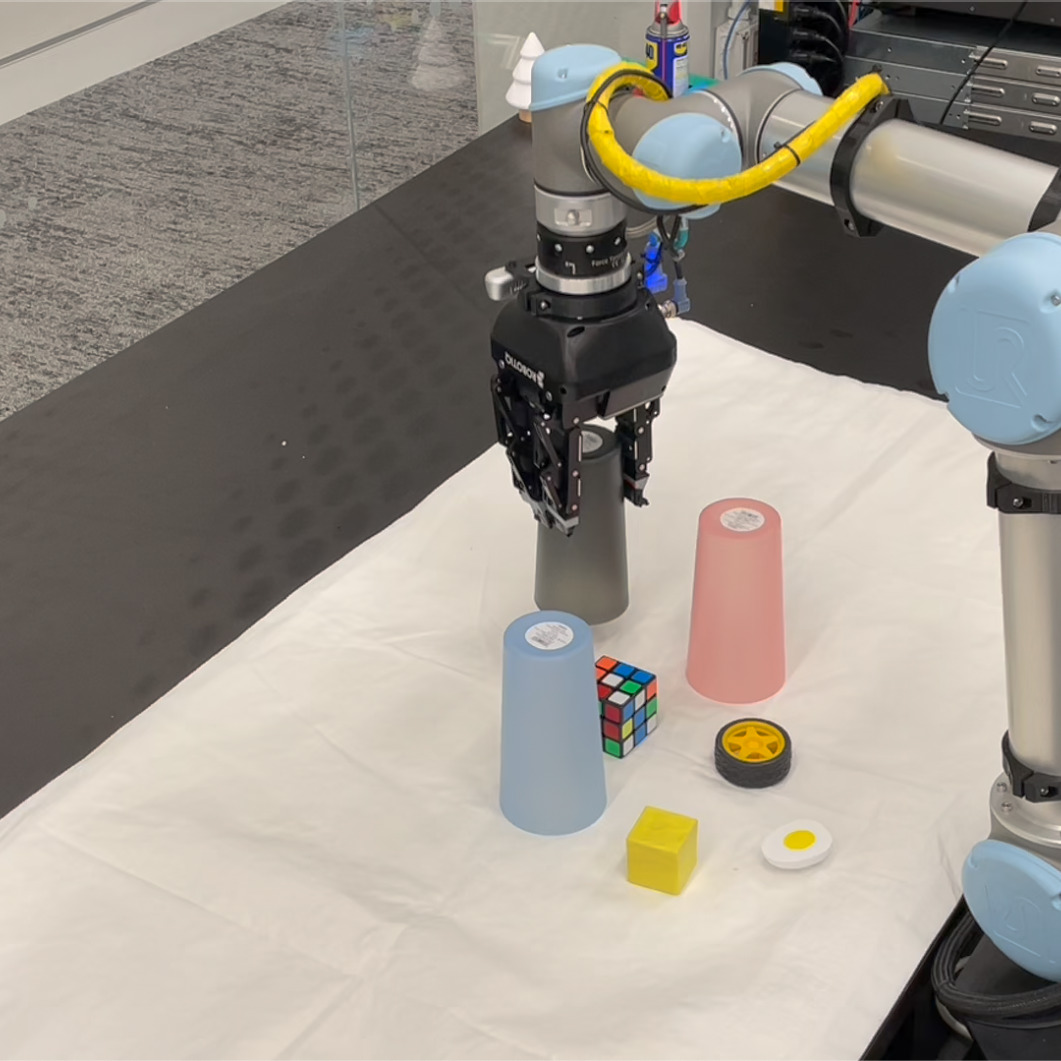}};%
        \node[empty,right = -1pt of statler2] (statler3) {\includegraphics[width=\mywidth]{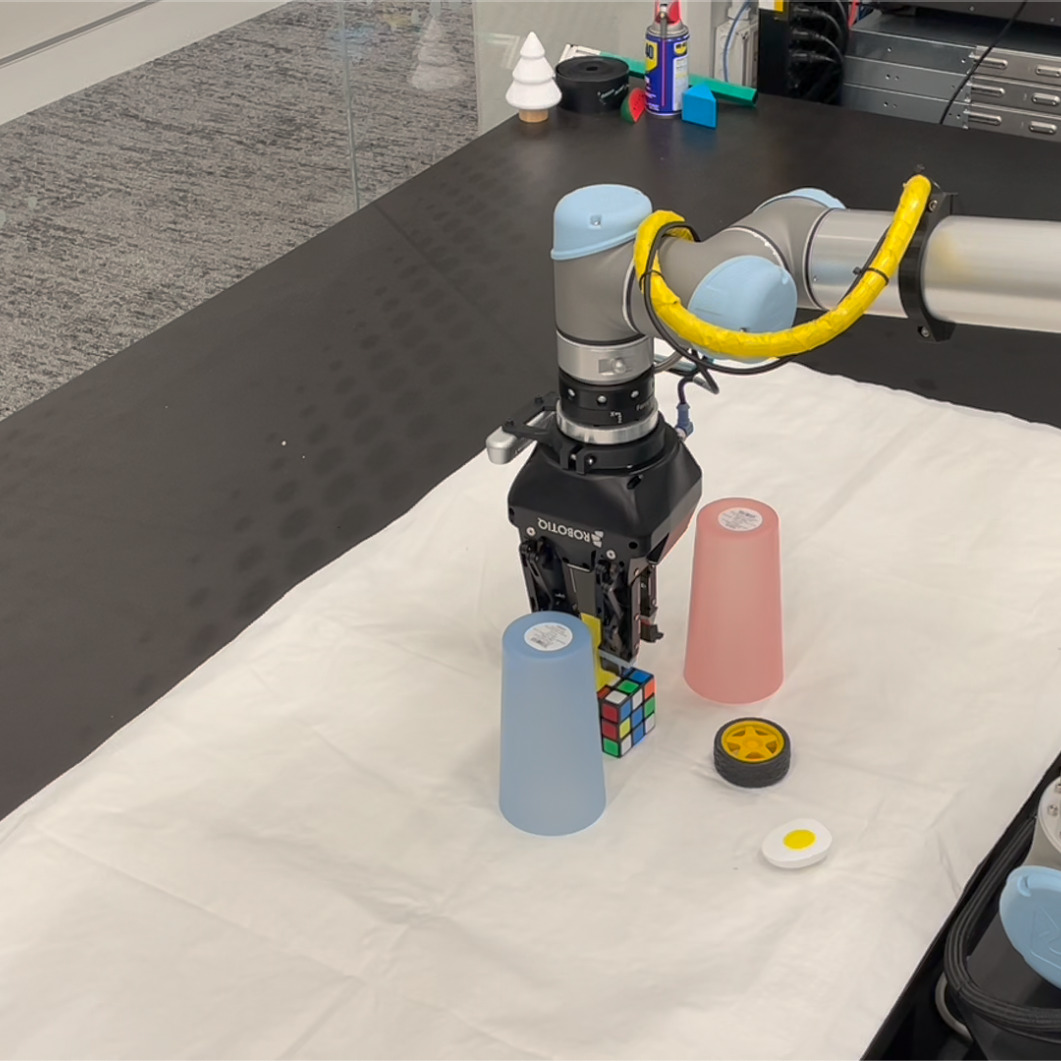}};%
        \node[empty,right = -1pt of statler3,draw=green, ultra thick] (statler4) {\includegraphics[width=\mywidth]{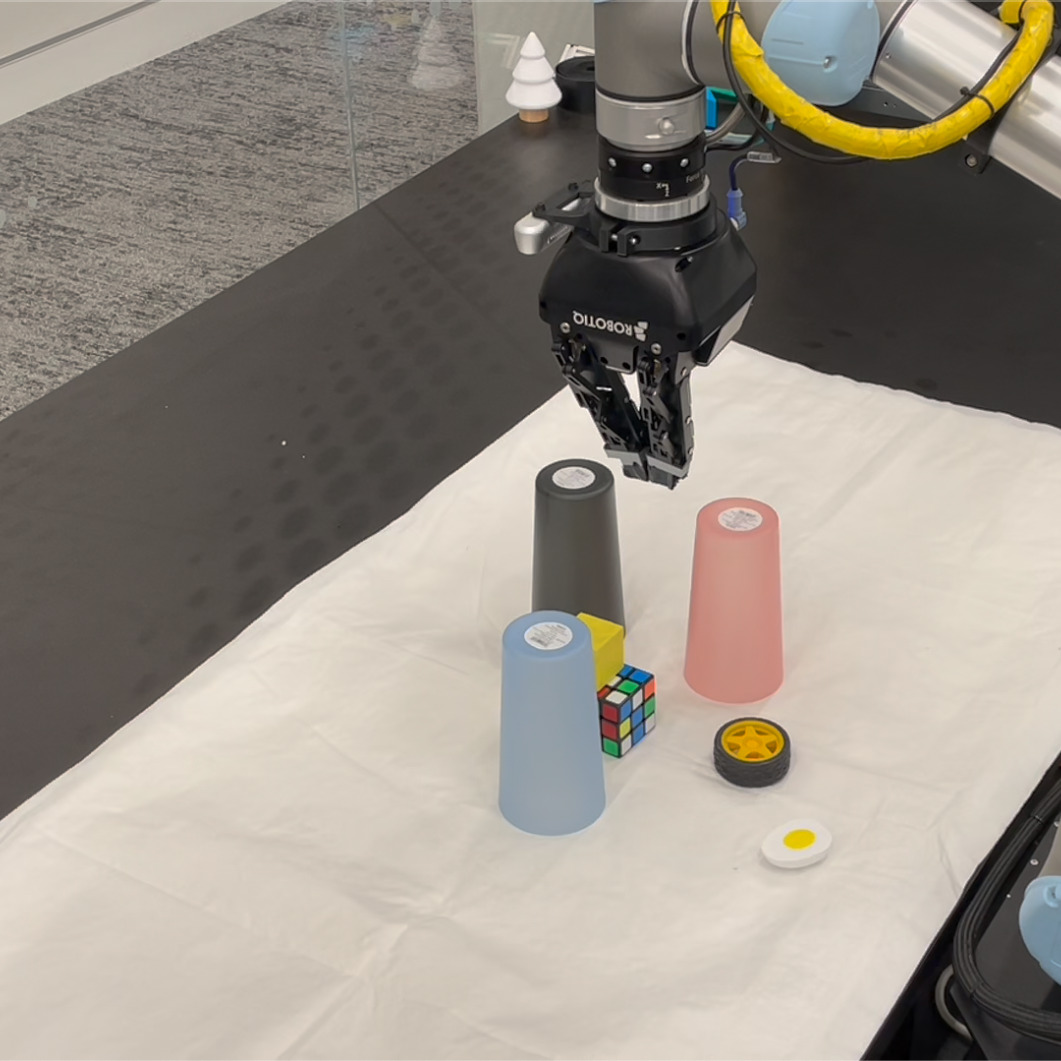}};%
        \node [circle,draw=black, fill=white, inner sep=1pt, minimum size=5pt, above right=5pt of statler4.south west, anchor=south west] (check) {\greencheck};
        \node[below = 2pt of statler0.south west] (line-start) {};
        \node[below = 2pt of statler4.south east] (line-end) {};
        \draw[thick,-stealth] (line-start) -- (line-end) node [midway,fill=white] {\footnotesize Time};
    \end{tikzpicture}
    \caption{A comparison of the resulting behavior for (top) \vanillallm and (bottom) our \acronym model for the real robot experiments for the given multi-sentence instruction. %
    Frames %
    correspond to instances when the robot has placed an object. In order to successfully carry out the instruction, the robot must subsequently remove the black cup immediately after using it to cover the yellow block so that it can place the yellow block on the Rubik's cube. However, the the baseline \vanillallm (top row, third frame) fails to move the black cup aside, leaving the yellow block covered. It then places a wrong object on top of the Rubik's cube.
    }\label{fig:real-robot-exp}
\end{figure}

In order to validate Statler on a real robot, we implement it on a UR5 arm in a similar tabletop domain as the simulated experiments. We use MDETR~\cite{kamath2021mdetr}, an open-vocabulary segmentation model, to obtain segmentation masks for objects within an RGB image captured by a RealSense camera mounted on the gripper. Using the predicted object mask and the depth image, the object point-cloud can be obtained, from which we estimate its center position and bounding box. All of the primitive functions are identical to those used in simulation. In this domain, the robot is asked to stack objects and cover objects with different colored cups. At any point, an object is only permitted to be covered by at most a single object or cover. If the robot is asked to manipulate the bottom object, it must put away the top one. If a new cover or object is to be stacked on it, the existing one must be removed.

We evaluate the performance of Statler vs.\ CaP in the real robot domain using $10$ episodes. Statler achieves episode and step (in parentheses) success rate of $\nfrac{40\%}{(70\%)}$, where $67\%$ of the failure cases are due to LLM reasoning while others are caused by either perception or manipulation issues. The success rate for CaP is $\nfrac{20\%}{(46\%)}$, where LLM reasoning accounts for $88\%$ of failures. In Figure~\ref{fig:real-robot-exp}, we also provide a short example where the CaP baseline fails. The difficulty is in recognizing that yellow block is hidden under the black cup, which must be removed before picking up the yellow block as \acronym correctly spots. Instead, CaP is not aware of this and tries to pick up the yellow block nonetheless, which leads MDETR to incorrectly detect the toy wheel that has yellow color in it as the yellow block.

\subsection{State-Maintenance Ablations}
To better understand \acronym's state-maintenance strategy, we consider three different approaches to tracking the state.

\begin{codefloat}
\begin{lstlisting}[caption={Portion of \acronym-Auto prompt.},label={lst:statler-auto},firstnumber=auto]
Your task is to maintain the status of these items using a JSON dictionary and update the status of the corresponding items after a new query.
This JSON dictionary will be commented, meaning that the starting character of each line is #.@
\end{lstlisting}
\end{codefloat}

The first (\acronym-Unified) employs a single LLM as both the world-state reader and writer using a prompt that interleaves \acronym's reader and writer prompts. At each step, the LLM first generates the action and then predicts the state that results from executing that action. The LLM then uses the resulting state when reasoning over the next query. Using a single LLM is conceptually simple, but it incurs an added burden for reasoning and generalization. Inspired by InstructGPT~\cite{ouyang2022training}, the second (\acronym-Auto) does not receive any in-context state-update examples for the world-state writer. Instead, we provide a natural language description of how the state should be maintained. Prompt~\ref{lst:statler-auto} shows the relevant portion of the prompt. With an instruction and no in-context state-update examples, we ran our model on the same set of tasks. The third (\acronym w/o State) ablates the world-state maintenance components of \acronym entirely, resulting in a model that reduces to Code-as-Policies.

\begin{table}[!ht]
    \centering
    \caption{Ablation episode (individual step) success rates.}\label{tb:ablations}
    \setlength{\tabcolsep}{2.5pt}
    \begin{tabularx}{1.0\linewidth}{lYYY}%
        \toprule
        & {Pick \& Place} & {Disinfection} & {Weight}\\
        \midrule
       \acronym w/o State & 
        \nfrac{0.00}{(0.54)} & %
        \nfrac{0.00}{(0.68)} & %
        \nfrac{0.00}{(0.84)} \\
       \acronym-Unified &
        \nfrac{0.40}{(0.85)} & %
        \nfrac{0.35}{(0.79)} & %
        \nfrac{0.50}{(0.92)} \\
       \acronym-Auto &
        \nfracb{0.75}{(0.88)} & %
        \nfracb{0.45}{(0.82)} & %
        \nfrac{0.40}{(0.90)} \\
        \midrule
       \acronym (ours) &
        \nfrac{0.50}{(0.88)} & %
        \nfrac{0.40}{(0.82)} & %
        \nfracb{0.55}{(0.93)} \\
       \bottomrule
    \end{tabularx}
\end{table}
Table~\ref{tb:ablations} compares the performance of \acronym to the three variations in terms of both their full-episode completion rates (using $20$ episodes for each domain) as well their individual step success rates. Without maintaining the world-state, \acronym w/o State fails to complete any episodes (recall that an episode is considered to be a failure if the model fails to respond to one of the user queries during the episode) and results in individual step success rates that are significantly lower than \acronym. Meanwhile, we see that \acronym's use of separate LLMs for the world-state reader and world-state writer results in consistently higher episode success rates compared with the use of a unified reader and writer (\acronym-Unified). The individual step success rates are higher for Pick-and-Place and Block Disinfection, and comparable for Relative Weight Reasoning. With regards to \acronym's use of separate LLMs for the world-state writer and reader, we note that in-context learning has been shown to be sensitive to variations in prompt templates, the order of examples, and the examples used
\cite{dong2022survey,zhao2021calibrate}. In light of this, it is plausible that the performance gains that we achieve by dividing our reader and writer may be attributed in part to this sensitivity, allowing the models to, in effect, become specialized at their respective tasks.  Interestingly, \acronym-Auto performs noticeably better than \acronym and \acronym-Unified with regards to the episode success rate on the Pick-and-Place and Block Disinfection domains, but comparable to \acronym in terms of the individual success rates, and worse for Relative Weight Reasoning.  \Ben{Do we want to comment on why auto performed better on episode success rate? This seems unaddressed\response{Peng} I cited InstructGPT}

\section{Related Work} \label{sec:relatid_works}

\textbf{Language Understanding for Robotics } A common approach for language understanding for robotic agents involves symbol grounding~\cite{harnad90}, whereby phrases are mapped to symbols in the robot's world model. Early work~\cite{winograd71,macmahon06} relies upon hand-engineered rules to perform this mapping. More recent methods replace these rules with statistical models the parameters of which are trained on annotated corpora~\cite{kollar10, matuszek10,
chen11, tellex11, matuszek12a, thomason15, howard14,
misra16, thomason16, thomason18, shridhar18, paul18}. Other methods use neural network-based architectures to jointly reason over natural language utterances and the agent's (visual) observations of the scene~\cite{mei2016listen, Anderson2017VisionandLanguageNI, fried18, zhu20, Min2021FILMFI}.

\textbf{LLMs for Robotics } Since LLMs are trained with Internet-scale corpora, their infused common sense have shown to help in the domain of robotics in terms of high-level planning from natural language instructions~\cite{Ahn2022DoAI,Liang2022CodeAP,huang2022language} for both object manipulation~\cite{wang2023programmatically,ren2023leveraging} and navigation tasks~\cite{majumdar2020improving,gadre2023cows,shah2023lm,Huang2022VisualLM}.
Combining LLMs with expressive visual-language embeddings also enables impressive capabilities \cite{Shridhar2021CLIPortWA}. This has led to efforts to push for general multi-modality embodied models \cite{Zeng2022SocraticMC, driess2023palm}. %

\textbf{Code Generation with LLMs } Code generation has been one of the most successful use cases for LLMs~\cite{DBLP:journals/corr/abs-2107-03374, Hendrycks2021MeasuringCC, Li2022CompetitionlevelCG, Chen2022CodeTCG, gpt3, DBLP:journals/corr/abs-2303-08774}. 
Since code can connect with executable APIs for tasks including computation, vision and manipulation, a large chunk of work has focused on code generation with different tools \cite{Schick2023ToolformerLM, Suris2023ViperGPTVI, Patil2023GorillaLL}. In particular, Code-as-policies~\cite{Liang2022CodeAP} is one of the first to use code generation within a robotics context.

\textbf{State Representation in Reasoning } 
The use of state representations have been shown to help in algorithmic reasoning tasks \cite{Nye2021ShowYW, Nam2022LearningTR}. Instead of using one forward pass to predict the execution result for the entire code snippet,  \citet{Nye2021ShowYW} proposes to spell out step-by-step intermediate outputs to help infer the final execution results. 
Also relevant are research efforts that aim to enhance language modeling by rolling out possible future tokens~\cite{du2023autoregressive}. 

\textbf{Language Models and Planning } Recent work shows that vanilla and instruction-tuned LLMs plan poorly \cite{valmeekam2023planning, silver2023generalized, liu2023llm+}. Some works propose using the LLM as an intermediary between natural language and a domain-specific programming language, and then uses a traditional planner~\cite{guan2023leveraging, liu2023llm+, wong2023word}. \citet{silver2023generalized} employ Chain-of-Thought and iterative reprompting with feedback on generated plans, but require GPT-4 for good performance. \citet{xiang2023language} use parameter-efficient finetuning of LLMs on top of traces from a world-model and show improved performance on related tasks.

\section{Conclusion}
\label{sec:conclusion}
We presented \acronym, a language model that maintains an explicit representation of state to support longer-horizon robot reasoning tasks. Integral to \acronym are a world-state reader that responds to a user query taking into account the current internal state, and a world-state writer that maintains the world state. 
Evaluations on various simulated and real robot manipulation tasks reveal that \acronym significantly outperforms contemporary models on  non-trivial tasks that require reasoning over the past. Ablations demonstrate the contributions of our world-state reader and writer, and suggest \acronym's flexibility to the state representation.%

\section{Acknowledgements}
We are grateful to National Science Foundation for enabling this work under HDR TRIPODS (No.\ 2216899), and to Adobe for supporting Hongyuan Mei through an Adobe Research gift.
We thank Luzhe Sun and Richard Xu for their help at the early stage of the project.

\flushcolsend

\bibliographystyle{IEEEtranN}
\bibliography{references}
\flushcolsend

\clearpage

\appendices

\section{Author Contributions}
\begin{itemize}
    \item \textbf{Takuma Yoneda} led the project. He discussed and came up with the seed idea with Jiading Fang. He also provided experiment designs from the early stages, and led the discussion and effort throughout the project.
    \item \textbf{Jiading Fang} initiated the hackathon that was the impetus to this project, shared the seed idea, contributed to writing.  %
    \item \textbf{Peng Li} implemented and conducted motivational experiments. He also designed and conducted the large part of our ablation experiments.
    \item \textbf{Huanyu Zhang} designed the evaluation episodes from the early stage of the project, and performed quantitative and qualitative analysis.  %
    \item \textbf{Tianchong Jiang} developed the PyBullet simulation environments for our experiments and created the corresponding  visualizations.
    \item \textbf{Shengjie Lin} led the real robot experiments with regards to task design, prompt development, perception integration, and evaluation execution.
    \item \textbf{Ben Picker} designed the evaluation episodes and helped with the partial automation of their generation and evaluation. Also contributed to research of related works and paper editing.
    \item \textbf{David Yunis} developed and executed the real robot experiments with Shengjie Lin. He also identified and compiled relevant works, and wrote most of the related work discussion as well as the real robot experiment section.
    \item \textbf{Hongyuan Mei} supervised the project. He also contributed significantly to restructuring the abstract and introduction.
    \item \textbf{Matthew R. Walter} supervised and guided the project from the initial hackathon through paper submission.
    He also helped Takuma to manage and distribute the workload between project members, which was critical to facilitate collaboration at this scale.
\end{itemize}

\section{Example episode sequences}
In this section, we show three example evaluation episodes (one per domain), to give a better idea of what \acronym is tasked to handle at test time.
The episodes are shown in Prompt \ref{lst:app-ep-pick-and-place}, \ref{lst:app-ep-disinfection} and \ref{lst:app-ep-rel-weight}. The \emph{temporal} queries that require temporal reasoning to answer correctly are highlighted in blue.
Python snippet under [Gold Code] shows the ``correct'' code the model is expected to generate. We note that the model can also output comments, \texttt{noop} and \texttt{update\_wm} functions (in the case of \acronym) that are not included in the gold code here.

The whole set of episodes ($20$ episodes per domain) can be found in our GitHub repository at \url{https://github.com/ripl/statler}.%

\begin{codefloat}
\begin{spacing}{0.975} 
\begin{lstlisting}[caption={Example evaluation episode (Pick-and-place domain). The queries in blue require temporal reasoning.}, label={lst:app-ep-pick-and-place},firstnumber=auto]
[Initial State]
# state = {
#     "objects": ["green block", "orange block", "white block", " cyan block", "golden bowl", "red bowl"],
#     "relations": [],
#     "green block": {},
#     "orange block": {},
#     "white block": {},
#     " cyan block": {},
#     "golden bowl": {},
#     "red bowl": {},
# }
[User Query 1] put the white block in the golden bowl
[Gold Code]
    put_first_on_second("white block", "golden bowl")
[User Query 2] put the cyan block on the green block
[Gold Code]
    put_first_on_second("cyan block", "green block")
[User Query 3] put the orange block in the empty bowl
[Gold Code]
    put_first_on_second("orange block", "red bowl")
@\textcolor{blue}{[User Query 4] put the block in the golden bowl on the block in the red bowl}@
[Gold Code]
    put_first_on_second("white block", "orange block")
@\textcolor{blue}{[User Query 5] is the green block in a higher position than the white block}@
[Gold Code]
    say("no")
[User Query 6] is the green block right above the orange block
[Gold Code]
    say("no")
\end{lstlisting}
\end{spacing}
\end{codefloat}

\begin{codefloat}
\begin{spacing}{0.975}
\begin{lstlisting}[caption={Example evaluation episode (Disinfection domain). The query in blue requires temporal reasoning.}, label={lst:app-ep-disinfection},firstnumber=auto]
[Initial State]
# state = {
#     "objects": ["teal block", "black block", "cyan block", "blue block", "tan bowl", "disinfector"], 
#     "relations": [], 
#     "disinfector": {"contains": []}, 
#     "teal block": {}, 
#     "black block": {}, 
#     "cyan block": {}, 
#     "blue block": {}, 
#     "tan bowl": {}
# }
[User Query 1] the teal block and the black block are dirty.
[User Query 2] The other blocks are clean
[User Query 3] Woops, somebody accidentally polluted the cyan block and the blue block
[User Query 4] Put the cyan block on the teal block
[Gold Code]
    put_first_on_second("cyan block", "teal block")
[User Query 5] Put the blue block and the black block in the disinfector
[Gold Code]
    put_first_on_second("blue block", "disinfector")
    put_first_on_second("black block", "disinfector")
[User Query 6] Put the blue block in the tan bowl
[Gold Code]
    put_first_on_second("blue block", "tan bowl")
[User Query 7] Put the blue block on the table.
[Gold Code]
    put_first_on_second("blue block", "table")
[User Query 8] Put the black block on the blue block
[Gold Code]
    put_first_on_second("black block", "blue block")
@\textcolor{blue}{[User Query 9] Put all the dirty blocks on the table.}@
[Gold Code]
    put_first_on_second("teal block", "table")
    put_first_on_second("cyan block", "table")
\end{lstlisting}
\end{spacing}
\end{codefloat}

\begin{codefloat}
\begin{spacing}{0.975} 
\begin{lstlisting}[caption={Example evaluation episode (Relative Weight domain). The query in blue requires temporal reasoning.}, label={lst:app-ep-rel-weight},firstnumber=auto]
[Initial State]
# state = {
#     "objects": ["black block", "orange block", green block", "red block", "gray bowl", "blue bowl"],
#     "relations": [],
#     "black block": {},
#     "orange block": {},
#     "green block": {},
#     "red block": {},
#     "gray bowl": {},
#     "blue bowl": {},
# }

[User Query 1] The black block is twice the weight of the green block
[User Query 2] Put the orange block in the gray bowl
[Gold Code]
    put_first_on_second("orange block", "gray bowl")
[User Query 3] The red block is twice the weight of the orange block
[User Query 4] The red block has the same weight of the black block
[User Query 5] Put the red block in the gray bowl
[Gold Code] 
    put_first_on_second("red block", "gray bowl")
@\textcolor{blue}{[User Query 6] Put blocks in the blue bowl so that their total weight becomes identical to what is in the gray bowl}@
[Gold Code]
    put_first_on_second("black block", "blue bowl")
    put_first_on_second("green block", "blue bowl")

\end{lstlisting}
\end{spacing}
\end{codefloat}

\section{Prompts}
For each domain, we provide tailored prompt that consists of example sequence of user queries and expected Python code.
To provide a concrete idea of our prompt design, we show our prompt of Code-as-Policies and \acronym for disinfection domain on Prompt \ref{lst:app-prompt-cap} and \ref{lst:app-prompt-statler}.

\begin{codefloat}
\begin{spacing}{0.975}
\begin{lstlisting}[caption={Prompt for Code-as-Policies (Disinfection domain)}, label={lst:app-prompt-cap},firstnumber=auto]
In the following, the robot deals with dirty and clean blocks.
A clean block becomes dirty when it touches another dirty block.
This includes when a dirty block is stacked on top of a clean block, and also when a dirty block is right under a clean block.
The table, bowls and the robot gripper are protected from any dirtiness, so they stay clean forever.
When a dirty block is put into the disinfector, it becomes clean immediately.

Instruction:
Aside from the built-in python functions and statements, the robot can only run the following functions:
`put_first_on_second`, `say` and `noop`.

Each code is carefully designed by professionals to meet all of these requirements.
===
# objects = ["cyan block", "yellow block", "brown block", "purple block", "blue block", "green bowl", "red bowl", "disinfector"]
# query: The cyan block and purple block are dirty
noop()
# query: The other blocks are clean
noop()
# query: Put the cyan block on the yellow block
put_first_on_second("cyan block", "yellow block")
# query: Put the brown block in the green bowl
put_first_on_second("brown block", "green bowl")
# query: Woops, somebody took out the brown block and dropped it on a dirty area
noop()
# query: Pick the cyan block and put it on the table
put_first_on_second("cyan block", "table")
# query: Move the yellow block into the disinfector
put_first_on_second("yellow block", "disinfector")
# query: Place all the clean blocks in the green bowl
# THINK: The clean blocks are yellow block and purple block
put_first_on_second("blue block", "green bowl")
put_first_on_second("yellow block", "green bowl")
# query: Put the cyan and purple block in the disinfector
put_first_on_second("cyan block", "disinfector")
put_first_on_second("purple block", "disinfector")
# query: Put the dirty blocks in the red bowl
# THINK: The only dirty block is the brown block
put_first_on_second("brown block", "red bowl")
# query: Pick the blue block and put it on the table
put_first_on_second("blue block", "table")
# query: Put the yellow block on the table
put_first_on_second("yellow block", "table")
# query: Put the brown block in the green bowl
put_first_on_second("cyan block", "green bowl")
# query: How many blocks are in the red bowl
# THINK: the red bowl is empty
say("There is no block in the red bowl")
===
\end{lstlisting}
\end{spacing}
\end{codefloat}

\begin{codefloat}
\begin{spacing}{0.975}
\begin{lstlisting}[caption={Prompt for \acronym (Disinfection domain). To avoid getting cluttered, many lines are omitted (denoted with ``omitted''). Please check our repository for full version.}, label={lst:app-prompt-statler},firstnumber=auto]
In the following, the robot deals with dirty and clean blocks.
A clean block becomes dirty when it touches another dirty block.
This includes when a dirty block is stacked on top of a clean block, and also when a dirty block is right under a clean block.
The table, bowls and the robot gripper are protected from any dirtiness, so they stay clean forever.
When a dirty block is put into the disinfector, it becomes clean immediately.

Instruction:
Aside from the built-in python functions and statements, the robot can only run the following functions:
`put_first_on_second`, `say` and `update_wm`.

Each code is carefully designed by professionals to meet all of these requirements.
===
# state = {
#     "objects": ["cyan block", "yellow block", "brown block", "purple block", "blue block", "green bowl", "red bowl", "disinfector"],
#     "relations": [],
#     "disinfector": {"contains": []},
#     "cyan block": {},
#     "yellow block": {},
#     "brown block": {},
#     "purple block": {},
#     "blue block": {},
#     "green bowl": {},
#     "red bowl": {}
# }
# query: The cyan block and purple block are dirty
update_wm("The cyan and purple blocks are dirty")
# state = { $\cdots$ omitted $\cdots$
# query: The other blocks are clean
update_wm("The brown block, yellow block, and blue block are clean")
# state = { $\cdots$ omitted $\cdots$
# query: Put the cyan block on the yellow block
put_first_on_second("cyan block", "yellow block")
update_wm("Put the cyan block on the yellow block.")
# state = {
#     "objects": ["cyan block", "yellow block", "brown block", "purple block", "blue block", "green bowl", "red bowl", "disinfector"],
#     "relations": [["cyan block is on yellow block"]],
#     "disinfector": {"contains": []},
#     "cyan block": {"is": ["dirty"]},
#     "yellow block": {"is": ["dirty"]},
#     "brown block": {"is": ["clean"]},
#     "purple block": {"is": ["dirty"]},
#     "blue block": {"is": ["clean"]},
#     "green bowl": {},
#     "red bowl": {}
# }
# query: Put the brown block in the green bowl
put_first_on_second("brown block", "green bowl")
update_wm("Put the brown block in the green bowl")
# state = { $\cdots$ omitted $\cdots$
# query: Woops, somebody took out the brown block and dropped it on a dirty area
update_wm("Put the brown block on the table, and make it dirty")
# state = { $\cdots$ omitted $\cdots$
# query: Pick the cyan block and put it on the table
put_first_on_second("cyan block", "table")
update_wm("Put the cyan block on the table")
# state = { $\cdots$ omitted $\cdots$
# query: Move the yellow block into the disinfector
put_first_on_second("yellow block", "disinfector")
update_wm("Put the yellow block in the disinfector")
# state = { $\cdots$ omitted $\cdots$
$\cdots\cdots\cdots\cdots\cdots\cdots\cdots$ omitted $\cdots\cdots\cdots\cdots\cdots\cdots\cdots$
# query: Put the brown block in the green bowl
put_first_on_second("cyan block", "green bowl")
update_wm("Put the cyan block in the green bowl")
# state = { $\cdots$ omitted $\cdots$
# query: How many blocks are in the red bowl
# THINK: the red bowl is empty
say("There is no block in the red bowl")
# state = { $\cdots$ omitted $\cdots$
===
\end{lstlisting}
\end{spacing}
\end{codefloat}

\section{Example failure cases}
We manually looked into examples that \acronym fails. Although providing insights on when and why the model fails would be helpful, we find that in many cases it is not trivial to interpret the model's mistakes.
In this section, we show three episodes (one per each domain) that the model failed, just to give some ideas on its failure modes.

\subsection{Pick-and-Place}
The episode with the generated state and code is shown in Prompt \ref{lst:llm-cot-12}. Given the fifth user query, "Is the green block in a higher position than the white block?" The current state correctly identifies the positions of both the white and green blocks (i.e., the white block is on the orange block, which is inside the red bowl, and the green block is on the table). However, with this information correctly absorbed as shown in its thinking process, the state reader still fails to recognize that the green block is in a lower position.

\begin{codefloat}
\begin{spacing}{0.975} 
\begin{lstlisting}[caption={Generated states and code on an evaluation episode (pick and place domain). Incorrect portions are highlighted in red.}, label={lst:llm-cot-12}, firstnumber=auto]
[Initial State]
# state = {
#     "objects": ["green block", "orange block", "white block", "cyan block", "golden bowl", "red bowl"],
#     "relations": [],
#     "green block": {},
#     "orange block": {},
#     "white block": {},
#     "cyan block": {},
#     "golden bowl": {},
#     "red bowl": {},
# }
$\cdots\cdots\cdots\cdots\cdots\cdots\cdots$ omitted $\cdots\cdots\cdots\cdots\cdots\cdots\cdots$
# state = {
#     "objects": ["green block", "orange block", "white block", "cyan block", "golden bowl", "red bowl"],
#      "relations": ["cyan block is on green block", "white block is on orange block"],
#      "green block": {},
#      "orange block": {},
#      "white block": {},
#      "cyan block": {},
#      "golden bowl": {},
#      "red bowl": {"contains": ["orange block"]},
# }
[User Query 5] is the green block in a higher position than the white block
[Generated Code]
# THINK: The cyan block is on the green block which is on the table
# THINK: The white block is on the orange block which is in the red bowl
@\sethlcolor{aired}\hl{\# THINK: Thus the green block is in a higher position}@
@\sethlcolor{aired}\hl{say("yes")}@

[Gold Code]
say("no (white block is on the orange block is in the red bowl, green block is on the table)")

$\cdots\cdots\cdots\cdots\cdots\cdots\cdots$ omitted $\cdots\cdots\cdots\cdots\cdots\cdots\cdots$
\end{lstlisting}
\end{spacing}
\end{codefloat}

\subsection{Disinfection}
The episode with generated state and code is on Prompt \ref{lst:llm-cot-13}. %
Given the eighth user query ``Put the black block on the blue block'', even though the current state correctly recognizes the both (black and blue) blocks to be \emph{clean}, generated next state marks them ``dirty'' by mistake (highlighted in red). This error by the state writer causes the failure of the code generation at the next step.
\begin{codefloat}
\begin{spacing}{0.975} 
\begin{lstlisting}[caption={Generated states and code on an evaluation episode (disinfection domain). Incorrect portions are highlighted in red.}, label={lst:llm-cot-13},firstnumber=auto]
[Initial State]
# state = {
#     "objects": ["teal block", "black block", "cyan block", "blue block", "tan bowl", "disinfector"], 
#     "relations": [], 
#     "disinfector": {"contains": []}, 
#     "teal block": {}, 
#     "black block": {}, 
#     "cyan block": {}, 
#     "blue block": {}, 
#     "tan bowl": {}
# }
$\cdots\cdots\cdots\cdots\cdots\cdots\cdots$ omitted $\cdots\cdots\cdots\cdots\cdots\cdots\cdots$
# state = {
#     "objects": ["teal block", "black block", "cyan block", "blue block", "tan bowl", "disinfector"], 
#     "relations": [["cyan block is on teal block"]], 
#     "disinfector": {"contains": ["black block"]}, 
#     "teal block": {"is": ["dirty"]}, 
#     "black block": {"is": ["clean"]}, 
#     "cyan block": {"is": ["dirty"]}, 
#     "blue block": {"is": ["clean"]}, 
#     "tan bowl": {}
# }
[User Query 8] Put the black block on the blue block
[Generated Code]
  put_first_on_second("black block", "blue block")
update_wm("Put the black block on the blue block")
[Gold Code]
  put_first_on_second("black block", "blue block")
[Generated State]
# state = {
#     "objects": ["teal block", "black block", "cyan block", "blue block", "tan bowl", "disinfector"], 
#     "relations": [["cyan block is on teal block"], ["black block is on blue block"]], 
#     "disinfector": {"contains": []}, 
#     "teal block": {"is": ["dirty"]}, 
#     "black block": {"is": ["@\sethlcolor{aired}\hl{dirty}@"]}, 
#     "cyan block": {"is": ["dirty"]}, 
#     "blue block": {"is": ["@\sethlcolor{aired}\hl{dirty}@"]}, 
#     "tan bowl": {}
# }
[User Query 9] Put all the dirty blocks on the table.
[Generated Code]
  put_first_on_second("teal block", "table")
  @\sethlcolor{aired}\hl{put\_first\_on\_second("black block", "table")}@
  put_first_on_second("cyan block", "table")
  @\sethlcolor{aired}\hl{put\_first\_on\_second("blue block", "table")}@
  update_wm("Put all the dirty blocks on the table")

[Gold Code]
  put_first_on_second("teal block", "table")
  put_first_on_second("cyan block", "table")
\end{lstlisting}
\end{spacing}
\end{codefloat}

\subsection{Relative weight reasoning}
The episode with generated state and code is on Prompt \ref{lst:llm-cot-14}. %
The model succeeds up to user query $5$. 
Given the next user query ``Put blocks in the blue bowl so that their total weight becomes identical to what is in the gray bowl'', the reasoning process that we can observe from \texttt{\# THINK:} comments on the generated code contains hallucination (highlighted in red). While the current state correctly identifies that the blue bowl is empty, the model \emph{thinks} the green block is in it, and as a result fails to move it over to the blue bowl.

\begin{codefloat}
\begin{spacing}{0.975} 
\begin{lstlisting}[caption={Generated states and code on an evaluation episode (relative weight reasoning domain). Incorrect portions are highlighted in red.}, label={lst:llm-cot-14},firstnumber=auto]
[Initial State]
# world_state = {
#     "objects": ["black block", "orange block", green block", "red block", "gray bowl", "blue bowl"],
#     "relations": [],
#     "black block": {},
#     "orange block": {},
#     "green block": {},
#     "red block": {},
#     "gray bowl": {},
#     "blue bowl": {},
# }
$\cdots\cdots\cdots\cdots\cdots\cdots\cdots$ omitted $\cdots\cdots\cdots\cdots\cdots\cdots\cdots$
[Generated State]
# state = {
#     "objects": ["black block", "orange block", green block", "red block", "gray bowl", "blue bowl"],
#     "relations": [],
#     "black block": {"weight": red_block.weight},
#     "orange block": {"weight": red_block.weight / 2},
#     "green block": {"weight": black_block.weight / 2},
#     "red block": {"weight": black_block.weight},
#     "gray bowl": {"contains": ["orange block", "red block"]},
#     "blue bowl": {},
# }
[User Query 6] Put blocks in the blue bowl so that their total weight becomes identical to what is in the gray bowl
[Generated Code]
  # THINK: In the gray bowl, we have orange block and red block
  @\sethlcolor{aired}\hl{\# THINK: In the blue bowl, we have green block}@
  # THINK: The orange block has half the weight of the red block
  # THINK: The green block has half the weight of the black block
  # THINK: We need to place a block that has `red_block.weight / 2`
  # THINK: The block that has the weight `red_block.weight / 2` is black block
  put_first_on_second("black block", "blue bowl")
  @\sethlcolor{aired}\hl{ }@
  update_wm("Put the black block in the blue bowl")
[Gold Code]
  put_first_on_second("black block", "blue bowl")
  put_first_on_second("green block", "blue bowl")
\end{lstlisting}
\end{spacing}
\end{codefloat}

\end{document}